# Polynomial Time and Private Learning of Unbounded Gaussian Mixture Models

Jamil Arbas[*]     Hassan Ashtiani[†]     Christopher Liaw[‡]

June 9, 2023

We study the problem of privately estimating the parameters of $d$-dimensional Gaussian Mixture Models (GMMs) with $k$ components. For this, we develop a technique to reduce the problem to its non-private counterpart. This allows us to privatize existing non-private algorithms in a blackbox manner, while incurring only a small overhead in the sample complexity and running time. As the main application of our framework, we develop an $(\varepsilon, \delta)$-differentially private algorithm to learn GMMs using the non-private algorithm of Moitra & Valiant [MV10] as a blackbox. Consequently, this gives the first sample complexity upper bound and first polynomial time algorithm for privately learning GMMs without any boundedness assumptions on the parameters. As part of our analysis, we prove a tight (up to a constant factor) lower bound on the total variation distance of high-dimensional Gaussians which can be of independent interest.

## 1 Introduction

The problem of learning the parameters of a Gaussian Mixture Model (GMM) is a fundamental problem in statistics, dating back to the early work of Pearson [Pea94] A GMM with $k$ components in $d$ dimensions can be represented as $(w_i, \mu_i, \Sigma_i)_{i=1}^{k}$, where $w_i$ is a mixing weight ($w_i \geq 0$, and $\sum_{i \in [k]} w_i = 1$), $\mu_i \in \mathbb{R}^d$ is a mean, and $\Sigma_i \in \mathbb{R}^{d \times d}$ is a covariance matrix (of the $i$-th Gaussian component). To draw a random instance from this GMM, one first samples an index $i \in [k]$ (with probability $w_i$) and then returns a random sample from the Gaussian distribution $\mathcal{N}(\mu_i, \Sigma_i)$. In this work we consider the problem of parameter estimation in the probably approximately correct (PAC) model, where the goal is to "approximately recover"[1] the parameters of an unknown GMM given only independent samples from it.

The sample complexity and computational complexity of learning the parameters of GMMs has been studied extensively. A breakthrough in this line of work was the development of polynomial-time methods (with respect to $d$) for learning GMMs with minimal separation requirements [MV10; BS10]. The running time and sample complexity of these methods is exponential $k$, which is generally necessary for parameter estimation [MV10].

The above approaches, however, may not maintain privacy of the individuals whose data has been used for the estimation. To address this issue, we adopt the rigorous and widely accepted notion of differential privacy (DP) [DMNS06]. At a high-level, DP ensures that the contribution of each individual has only a small (indistinguishable) effect on the output of the estimator. The classical notion of $\varepsilon$-DP (pure DP) is, however, quite restrictive. For instance, even estimating the mean of an unbounded univariate Gaussian random variable in this model is impossible. Therefore, in line with recent work on private estimation in unbounded domains, we consider the $(\varepsilon, \delta)$-DP (i.e. approximate differential privacy [DKMMN06]) model.

---

[*]McMaster University, `arbasj@mcmaster.ca`.
[†]McMaster University, `zokaeiam@mcmaster.ca`. Hassan Ashtiani is also a faculty affiliate at Vector institue.
[‡]Google, `cvliaw@google.com`

[1]See Definition 1.4 for the precise notion of distance.

For the simpler case of multivariate Gaussians (without any boundedness assumptions on the parameters), it has been shown that learning with a finite number of samples is possible in the $(\varepsilon, \delta)$-DP model [AAK21]. More recently, computationally efficient estimators have been devised for the same task [AL22; KMSSU22; KMV22]. This begs answering the corresponding question for GMMs.

> Is there an $(\varepsilon, \delta)$-DP estimator with a bounded sample complexity for learning unbounded GMMs?
> Is there a polynomial time estimator (in terms of $d$) for the same task?

Note that if additional boundedness[2] and strong separation[3] assumptions are made about the GMM, then the work of Kamath et al. [KSSU19] offers a positive answer to the above question in the $\varepsilon$-DP model. It has also been shown that the separation condition between the components can be significantly weakened [CKMST21; CCdEIST23]. Our aim is, however, learning *unbounded* GMMs with *minimal* separation assumptions.

To approach this problem, it is natural to ask if there is a general reduction from the private learning of GMMs to its non-private counterpart. If so, this would enable us to easily reuse existing results for non-private learning of GMMs.

> Is there a reduction from private to non-private learning of GMMs that incurs only a polynomial time and polynomial sample overhead?

The main result of this paper is the existence of such a reduction; see Theorem 6.2 for a rigorous version.

**Theorem 1.1** (**Private to Non-private Reduction for GMMs, Informal**). *There is a reduction from learning the parameters of a GMM in the $(\varepsilon, \delta)$-DP model to its non-private counterpart. Moreover, this reduction adds only polynomial time and sample overhead in terms of $d$ and $k$.*

This reduction, along with the non-private learner of Moitra & Valiant [MV10] gives the first finite sample complexity upper bound for learning the parameters of unbounded GMMs in the $(\varepsilon, \delta)$-DP model. Moreover, the resulting estimator essentially inherits all the properties of the non-private estimator of Moitra & Valiant [MV10]; it runs in time that is polynomial in $d$ (for fixed $k$) and shares the advantage of requiring provably minimal separability assumptions on the components of the GMM. We refer the reader to the related work section for a comparison with Cohen et al. [CKMST21] and Chen et al. [CCdEIST23].

## 1.1 Related Work

**Private Learning of a Single Gaussian.** Karwa & Vadhan [KV17] established polynomial time and sample efficient methods for learning the mean and variance of a univariate Gaussian in both the pure and approximate-DP settings. Namely, in the $(\varepsilon, \delta)$-DP setting, they can recover the mean and variance of the Gaussian without any boundedness assumption on the parameters. This result can be generalized to the multivariate setting [KLSU19; BDKU20], where one finds Gaussians that approximate the underlying Gaussian in terms of total variation distance. However, the sample complexity of these methods depends on the condition number of the covariance matrix, and requires a priori bounds on the range of the parameters. The first finite sample complexity bound for privately learning unbounded Gaussians appeared in [AAK21], nearly matching the sample complexity lower bound of [KMS22]. The work of [AAK21] relies on a private version of the minimum distance estimator [Yat85] and is based on ideas from the private hypothesis selection method [BSKW19]. However, this method is not computationally efficient. Recently, several papers offered $(\varepsilon, \delta)$-DP and computationally efficient algorithms for learning unbounded Gaussians [AL22; KMSSU22; KMV22], where the work of Ashtiani & Liaw [AL22] achieved a near-optimal sample complexity for this task. Part of the approach of Ashtiani & Liaw [AL22] is a sub-sample-and-aggregate scheme which we modify and use in this paper. FriendlyCore [TCKMS22] is an alternative sample-and-aggregate framework that can be used for privately learning unbounded Gaussians. It is noteworthy that the approaches of [AL22; KMV22] work in the robust setting as well albeit with sub-optimal sample complexities. The recent work of Alabi et al. [AKTVZ22] offers a robust and private learner with near-optimal sample requirements in terms of dimension. Finally, Hopkins et al. [HKMN23] ticks all the boxes by offering a sample near-optimal, robust, and efficient learner for unbounded Gaussians.

---

[2]They assume there are known quantities $R, \sigma_{max}, \sigma_{min}$ such that $\forall i \in [k], \|\mu_i\|_2 \leq R$ and $\sigma_{min}^2 \leq \|\Sigma_i\| \leq \sigma_{max}^2$.
[3]They assume $\forall i \neq j, \|\mu_i - \mu_j\|_2 \geq \widetilde{\Omega}\left(\sqrt{k} + \sqrt{\frac{1}{w_i} + \frac{1}{w_j}}\right) \cdot \max\left\{\|\Sigma_i^{1/2}\|, \|\Sigma_j^{1/2}\|\right\}$.

Another related result is a sample-efficient and computationally efficient method for learning bounded and high-dimensional Gaussians in the $\varepsilon$-DP model [HKM22]. There is also work on the problem of private mean estimation with respect to Mahalanobis distance [BGSUZ21; DHK23]. Finding private and robust estimators [LKKO21] and also the interplay between robustness and privacy [DL09; GH22; LKO22; HKMN23; AUZ23] are subjects of a few recent papers.

**Parameter Learning for GMMs with PAC Guarantees.** Given i.i.d. samples from a GMM, can we approximately recover its parameters? There has been an extensive amount of research in developing sample efficient and computationally efficient methods for learning the parameters of a GMM [Das99; SK01; VW04; AM05; BV08; KMV10; FSO06; BS09; HP14; HK13; ABGRV14; RV17; KSS18; HL18; LL22]. Remarkably, Moitra & Valiant [MV10] and Belkin & Sinha [BS10] presented the first polynomial time algorithms for learning general GMMs with unbounded components and under minimal separation assumptions. Here, the focus is on designing polynomial methods *with respect to dimension*, since having an exponential dependence on the number of components is inevitable [MV10] (unless the components are well-separated). These results haven been recently extended to the robust setting [BDJKKV22; LM21; LM22].

In the private setting, the early work of Nissim, Raskhodnikova & Smith [NRS07] offered an $(\varepsilon, \delta)$-DP estimator for the means of a GMM, in the special case where the components share the same mixing weight and the same (known) covariance matrix. Note that this result also inherits the strong separation assumption (of $\Omega(k^{1/4})$) between the Gaussian components from Vempala & Wang [VW04]. The recent (and independent) work of Chen et al. [CCdEIST23] shows that this separation can be significantly relaxed. These results are, however, for learning (unbounded) spherical Gaussian mixtures and unlike our work, they do not learn the covariance matrices.

Perhaps more related to our paper is the work of Kamath et al. [KSSU19], which offers an $(\varepsilon, \delta)$-DP parameter learning method for GMMs with unknown mixing weights, means, and covariance matrices. In fact, their approach is a privatized version of Achlioptas & McSherry [AM05]. However, their method only works when the parameters of the Gaussian components are bounded and the means are strongly separated (i.e., $\Omega(\sqrt{k})$-separated). In a related work, Bie, Kamath & Singhal [BKS22] show how can one use public data to improve this approach.

Finally, the more recent work of Cohen et al. [CKMST21] improves over Kamath et al. [KSSU19] by offering a better sample complexity and requiring weaker separation between the clusters. They show we can learn GMMs privately if we are given *(i)* a private learner for Gaussians and *(ii)* a non-private clustering method (i.e., an algorithm that can label the data points accurately based on their clusters). Given the generality of their reduction, one can plug a non-private clustering method that requires weaker separation between the components. On top of the separation requirements for the clustering method, their result also requires an $\Omega(\log n)$ separation between the means, where $n$ is the number of samples. Given, that $n$ is polynomial in $1/\varepsilon, k$ and $d$, their approach requires mild separation (i.e., logarithmic in these parameters).

Unlike Kamath et al. [KSSU19] and Cohen et al. [CKMST21], our approach does not require a priory bounds on the range of the Gaussian means or on the condition numbers of the covariance matrices. It may be possible to extend the result of Cohen et al. [CKMST21] to the unbounded setting, e.g., using the private Gaussian estimator of Ashtiani & Liaw [AL22]; yet, there are some subtle challenges for clustering when the condition number of Gaussian components are high. Another difference lies in the separation requirements. While Cohen et al. [CKMST21] relaxes the separation requirements of Kamath et al. [KSSU19], it still requires a (mild) separation of $\log(kd/\varepsilon)$ between the components. Moreover, additional separation requirements must be met for the non-private clustering method to work. While for spherical Gaussians this requirement is rather weak (i.e., $\sqrt{\log k}$ separation [LL22], we are not aware of clustering methods that work for non-spherical Gaussians and require weak separation (e.g., independent of the condition number of the Gaussian components). In contrast, our approach uses non-private parameter estimation (rather than clustering) and requires only "minimal separation" that is independent of $\epsilon, k, d$ (see Definition 6.4).

**Density Estimation for GMMs.** In the density estimation problem for GMMs, the goal is to recover a distribution that is close (often in total variation distance) to the underlying GMM. From the statistical point of view, the sample complexity of this problem has been settled up to logarithmic factors [DL01; ABM18; ABHLMP18; ABHLMP20]. Unlike the parameter learning setting, the sample complexity is actually polynomial both in terms of the dimension and the number of components. There are also computationally efficient algorithms for learning one-dimensional GMMs [CDSS14; ADLS17; LLM22; WX18; LS17]. Designing a polynomial time (with respect to dimension and number of components) algorithm for learning GMMs with respect to total variation distance remains an important open problem. Solving this problem is challenging as it requires overcoming known statistical query lower bounds for the problem [DKS17].

In the private setting, one can use the private hypothesis selection framework Bun et al. [BSKW19] or the private minimum distance estimator [AAK21] to learn classes that admit a finite cover. Therefore, GMMs with bounded parameters admit an $\varepsilon$-DP finite sample complexity guarantee [BSKW19]. A polynomial sample complexity upper bound is known for learning axis-aligned GMMs in the $(\varepsilon, \delta)$-DP model *without any boundedness assumptions on the parameters* [AAL21]. Extending this results to general GMMs remains an open problem. Furthermore, designing private and computationally efficient estimators for GMMs remains open even in the one dimensional setting. Another relevant result is a lower bound on the sample complexity of learning GMMs with known covariance matrices [ASZ21].

## 1.2 Preliminaries

We use $\|v\|_2$ to denote the Euclidean norm of a vector $v \in \mathbb{R}^d$ and $\|A\|_F$ (resp. $\|A\|$) to denote the Frobenius (resp. spectral) norm of a matrix $A \in \mathbb{R}^{d \times d}$.

In this paper, we write $\mathcal{S}^d$ to denote the positive-definite cone in $\mathbb{R}^{d \times d}$. Let $\mathcal{G}(d) = \{\mathcal{N}(\mu, \Sigma) : \mu \in \mathbb{R}^d, \Sigma \in \mathcal{S}^d\}$ be the family of $d$-dimensional Gaussians. We can now define the class $\mathcal{G}(d, k)$ of mixtures of Gaussians as follows.

**Definition 1.2** (Gaussian Mixtures). The class of mixtures of $k$ Guassians in $\mathbb{R}^d$ is defined by $\mathcal{G}(d, k) := \left\{ \sum_{i=1}^{k} w_i G_i : G_i \in \mathcal{G}(d), w_i \geq 0, \sum_{i=1}^{k} w_i = 1 \right\}$.

We represent the Gaussian Mixture Model (GMM) by a set of $k$ tuples $(w_i, \mu_i, \Sigma_i)_{i=1}^{k}$, where each tuple represents the mean, covariance matrix, and mixing weight of one of its components. Note that the order of the components is important in our notation, since the order of the output may have an impact on the privacy.

In the following definition and the remainder of the paper, we may abuse terminology and refer to a distribution via its probability density function (p.d.f.).

**Definition 1.3** (Total Variation Distance). Given two absolutely continuous probability measures $f(x), g(x)$ on $\mathbb{R}^d$, the total variation (TV) distance between $f$ and $g$ is defined as $d_{\text{TV}}(f(x), g(x)) = \frac{1}{2} \int_{\mathbb{R}^d} |f(x) - g(x)| \, dx$.

A standard way to define the distance between two GMMs is as follows ([MV10], Definition 2).

**Definition 1.4** (The Distance between Two GMMs). The $\text{dist}_{\text{GMM}}$ distance between two GMMs is defined by

$$\text{dist}_{\text{GMM}}\left((w_i, \mu_i, \Sigma_i)_{i=1}^k, (w_i', \mu_i', \Sigma_i')_{i=1}^k\right) = \min_\pi \max_{i \in [k]} \max\left\{|w_i - w'_{\pi(i)}|, d_{\text{TV}}\left(\mathcal{N}(\mu_i, \Sigma_i), \mathcal{N}(\mu'_{\pi(i)}, \Sigma'_{\pi(i)})\right)\right\}$$

where $\pi$ is chosen from the set of all permutations over $[k]$.

If $X$ (resp. $Y$) is a random variable distributed according to $f$ (resp. $g$), we write $d_{\text{TV}}(X, Y) = d_{\text{TV}}(f, g)$. We drop the reference to the p.d.f. of the random variable when it is clear or implicit from context.

## 1.3 Differential Privacy Basics

At a high-level, an algorithm is differentially private if, given two datasets that differ only in a single element, the output distribution of the algorithm are nearly the same[4].

**Definition 1.5** (Neighbouring Datasets). Let $\mathcal{X}, \mathcal{Y}$ denote sets and $n \in \mathbb{N}$. Two datasets $D = (X_1, \ldots, X_n), D' = (X_1, \ldots, X_n) \in \mathcal{X}^n$ are said to be *neighbouring* if $d_H(D, D') \leq 1$ where $d_H$ denotes Hamming distance, i.e., $d_H(D, D') = |\{i \in [n] : X_i \neq X_i'\}|$.

**Definition 1.6** (($\varepsilon, \delta$)-Indistinguishable). Let $D, D'$ be two distributions defined on a set $\mathcal{Y}$. Then $D, D'$ are said to be ($\varepsilon, \delta$)-indistinguishable if for all measurable $S \subseteq \mathcal{Y}$, $\mathbb{P}_{Y \sim D}[Y \in S] \leq e^\varepsilon \mathbb{P}_{Y \sim D'}[Y \in S] + \delta$ and $\mathbb{P}_{Y \sim D'}[Y \in S] \leq e^\varepsilon \mathbb{P}_{Y \sim D}[Y \in S] + \delta$.

**Definition 1.7** (($\varepsilon, \delta$)-Differential Privacy [DMNS06]). A randomized mechanism $\mathcal{M} \colon \mathcal{X}^n \to \mathcal{Y}$ is said to be ($\varepsilon, \delta$)-differentially private if for all neighbouring datasets $D, D' \in \mathcal{X}^n$, $\mathcal{M}(D)$ and $\mathcal{M}(D')$ are ($\varepsilon, \delta$)-indistinguishable.

## 1.4 Techniques

The techniques in this paper are inspired by the techniques in Ashtiani & Liaw [AL22] which are based on the Propose-Test-Release framework [DL09] and the Subsample-And-Aggregate framework [NRS07]. Given a dataset $D$, we first split $D$ into $t$ sub-datasets and run a non-private algorithm $\mathcal{A}$ on each of the sub-datasets. Next, we privately check if most of the outputs of $\mathcal{A}$ are "well-clustered" (i.e., are close to each other). If not, then the algorithm fails as this suggests that the outputs of the non-private algorithm are not very stable (either due to lack of data or simply that the non-private algorithm is sensitive to its input). On the other hand, if most of the outputs are well-clustered then we can aggregate these clustered outputs and release a noisy version of it. There are, however, multiple additional technical challenges that need to be addressed.

One core difficulty is the issue of the ordering of the Gaussian components. Namely, the non-private GMM learners may output GMM components in different orders. Therefore, aggregating these non-private solutions (e.g., by taking their weighted average in the style of Ashtiani & Liaw [AL22] seems impossible. We therefore propose to skip the aggregation step all together by simply picking an arbitrary solution from the cluster. Therefore, our private populous estimator (PPE) simplifies and generalizes the private populous mean estimator (PPME) framework of Ashtiani & Liaw [AL22], making it applicable to general semimetric spaces (and therefore GMMs). A precise discussion of this framework is presented in Subsection 2.1.

---

[4]For sake of simplicity, we consider data sets to be ordered and therefore the neighboring data sets are defined based on their Hamming distances. However, one can easily translate guarantees proven for the ordered setting to the unordered one; see Proposition D.6 in [BGSUZ21].

Another challenge is designing an appropriate mechanism for adding noise to GMMs. As discussed above, our framework requires that we are able to release a noisy output of a candidate output. More precisely, given two neighbouring datasets $Y_1, Y_2$, we want to design a mechanism $\mathcal{B}$ such that $\mathcal{B}(Y_1), \mathcal{B}(Y_2)$ are indistinguishable whenever $Y_1, Y_2$ are sufficiently close. As in Ashtiani & Liaw [AL22], we refer to such a mechanism as a "masking mechanism". In the context of mixture distributions with $k$ components, a candidate output corresponds to a $k$-tuple where each element of the tuple contain the parameters and the mixing weight of a single component. We prove that, if one can design a masking mechanism for a *single* component then it is possible to use this masking mechanism as a blackbox to design a masking mechanism for the $k$-tuple with only a poly($k$) overhead in the running time. One important ingredient is that we randomly shuffle the components, making the output invariant to the order of the components.

Another challenge related to the order of components is that computing the distance between two GMMs based on Definition 1.4 requires minimizing over all permutations. A naive method for computing this distance could require exponential time but we show this task can be done in polynomial time using a simple reduction to bipartite matching.

To showcase the utility of the above framework, we show that it is straightforward to apply the framework to privately learning mixtures of Gaussians. We design a masking mechanism of a single Gaussian component which consists of mixing the weight, the mean, and the covariance matrix. Masking the mixing weight is fairly standard while masking the mean and the covariance matrix can be done using known results (e.g. by using [AL22], Lemma 5.2) for the covariance matrix and a similar technique for the mean).

Finally, we note that, in some of the literature for Gaussian mixtures, the results usually assert that for each Gaussian component $\mathcal{N}(\mu, \Sigma)$, the algorithm returns $\hat{\mu}, \hat{\Sigma}$ such that $\mathcal{N}(\mu, \Sigma)$ and $\mathcal{N}(\hat{\mu}, \hat{\Sigma})$ are close in *total variation* distance (e.g. [MV10]). Our framework requires that $\hat{\mu}$ (resp. $\hat{\Sigma}$) is close to $\mu$ (resp. $\Sigma$) for some appropriate norm. Intuitively, this ought to be the case but no tight characterization was previously known unless the Gaussians had the same mean ([DMR18], Theorem 1.1). In this paper, we prove the following tight characterization between the TV distance of a Gaussian and its parameters. We believe that such a result may be of independent interest.

**Theorem 1.8.** *Let $\mu_1, \mu_2 \in \mathbb{R}^d$ and $\Sigma_1, \Sigma_2$ be $d \times d$ positive-definite matrices. Suppose that we have $d_{\mathrm{TV}}(\mathcal{N}(\mu_1, \Sigma_1), \mathcal{N}(\mu_2, \Sigma_2)) < \frac{1}{600}$. Let*

$$\Delta = \max\left\{\|\Sigma_1^{-1/2}\Sigma_2\Sigma_1^{-1/2} - I_d\|_F, \|\Sigma_1^{-1/2}(\mu_1 - \mu_2)\|_2\right\}.$$

*Then*

$$\frac{1}{200}\Delta \leq d_{\mathrm{TV}}\left(\mathcal{N}(\mu_1, \Sigma_1), \mathcal{N}(\mu_2, \Sigma_2)\right) \leq \frac{1}{\sqrt{2}}\Delta.$$

**Remark 1.9.** *Note that the total variation distance between $\mathcal{N}(\mu_1, \Sigma_1)$ and $\mathcal{N}(\mu_2, \Sigma_2)$ is symmetric (i.e. swapping $\mu_1, \Sigma_1$ and $\mu_2, \Sigma_2$ does not affect the total variation distance) while the definition of $\Delta$ is not. Thus, the theorem can be automatically strengthened to show that the bounds also hold when $\Delta$ is defined by swapping $\Sigma_1$ and $\Sigma_2$.*

## 2 Private Populous Estimator

In this section, we describe our main framework which we call the "private populous estimator" (PPE). Before that, we need a few definitions.

**Semimetric spaces.** In our application, we need to deal with distance functions which only satisfy an *approximate* triangle inequality that hold only when the points are sufficiently close together. To that end, we first define the notion of a semimetric space.

**Definition 2.1** (Semimetric Space). We say $(\mathcal{F}, \text{dist})$ is a semimetric space if for every $F, F_1, F_2, F_3 \in \mathcal{F}$, the following conditions hold.

1. **Non-negativity.** $\text{dist}(F, F) = 0$; $\text{dist}(F_1, F_2) \geq 0$.
2. **Symmetry.** $\text{dist}(F_1, F_2) = \text{dist}(F_2, F_1)$.
3. **$z$-approximate $r$-restricted triangle inequality.** Let $r > 0$ and $z \geq 1$. If $\text{dist}(F_1, F_2), \text{dist}(F_2, F_3) \leq r$ then $\text{dist}(F_1, F_3) \leq z \cdot (\text{dist}(F_1, F_2) + \text{dist}(F_2, F_3))$.

**Masking mechanism.** Intuitively, a masking mechanism $\mathcal{B}$ is a random function that returns a noisy version of its input, with the goal of making close inputs indistinguishable. Formally, we define a masking mechanism as follows.

**Definition 2.2** (Masking Mechanism ([AL22], Definition 3.3)). Let $(\mathcal{F}, \text{dist})$ be a semimetric space. A randomized function $\mathcal{B}\colon \mathcal{F} \to \mathcal{F}$ is a $(\gamma, \varepsilon, \delta)$-masking mechanism for $(\mathcal{F}, \text{dist})$ if for all $F, F' \in \mathcal{F}$ satisfying $\text{dist}(F, F') \leq \gamma$, we have that $\mathcal{B}(F), \mathcal{B}(F')$ are $(\varepsilon, \delta)$-indistinguishable. Further, $\mathcal{B}$ is said to be $(\alpha, \beta)$-concentrated if for all $F \in \mathcal{F}$, $\mathbb{P}[\text{dist}(\mathcal{B}(F), F) > \alpha] \leq \beta$.

## 2.1 The Private Populous Estimator (PPE)

In this section, we define the PPE framework which allows us to use non-private algorithms to design private algorithms. We represent the non-private algorithm by $\mathcal{A}\colon \mathcal{X}^* \to \mathcal{Y}$ which takes elements from a dataset as inputs and outputs an element in $\mathcal{Y}$. PPE requires two assumptions. Firstly, we assume that $(\mathcal{Y}, \text{dist})$ is a semimetric space. Secondly, we assume that we have access to an efficient masking mechanism for $(\mathcal{Y}, \text{dist})$.

The PPE framework we introduce in this section can be seen as a somewhat generalized version of the framework used in Ashtiani & Liaw [AL22] and requires fewer assumptions. Given a dataset $D$ as inputs, we partition $D$ into $t$ disjoint subsets. Next, we run the non-private algorithm $\mathcal{A}$ on each of these subsets to produce $t$ outputs $Y_1, \ldots, Y_t$. We then privately check if most of the $t$ outputs are close to each other. If not, PPE fails. Otherwise, it chooses a $Y_j$ that is close to more than 60% of other $Y_i$'s. It then adds noise to $Y_j$ using a masking mechanism $\mathcal{B}$, and returns the masked version of $Y_j$. The formal details of the algorithm can be found in Algorithm 1.

---
**Algorithm 1** Private Populous Estimator

**Input:** Dataset $D = (X_1, \ldots, X_m)$, any algorithm $\mathcal{A}\colon \mathcal{X}^* \to \mathcal{Y}$, parameters $r, \varepsilon, \delta > 0, z \geq 1, t \in \mathbb{N}_{\geq 1}$.

1: Let $s \leftarrow \lfloor m/t \rfloor$.
2: For $i \in [t]$, let $Y_i \leftarrow \mathcal{A}(\{X_\ell\}_{\ell=(i-1)s+1}^{is})$.
3: For $i \in [t]$, let $q_i \leftarrow \frac{1}{t}|\{j \in [t] : \text{dist}(Y_i, Y_j) \leq r/2z\}|$.
4: Let $Q \leftarrow \frac{1}{t}\sum_{i \in [t]} q_i$.
5: Let $Z \sim \text{TLap}(2/t, \varepsilon, \delta)$.
6: Let $\widetilde{Q} \leftarrow Q + Z$.
7: If $\widetilde{Q} < 0.8 + \frac{2}{t\varepsilon} \ln\left(1 + \frac{e^\varepsilon - 1}{2\delta}\right)$, fail and return $\bot$.
8: $j = \min\{i : q_i > 0.6\}$.
9: Return $\widetilde{Y} = \mathcal{B}(Y_j)$.

---

The following theorem establishes the privacy and accuracy of Algorithm 1. The proof can be found in Appendix D.1.

**Theorem 2.3.** *Suppose that $(\mathcal{Y}, \mathrm{dist})$ satisfies a $z$-approximate $r$-restricted triangle inequality. Further, suppose that $\mathcal{B}$ is a $(r, \varepsilon, \delta)$-masking mechanism.*

- **Privacy.** *For $t > 5$, Algorithm 1 is $(2\varepsilon, 4e^\varepsilon \delta)$-DP.*
- **Utility.** *Suppose $\alpha \le r/2z$ and $t \ge \left(\frac{20}{\varepsilon} \ln\left(1 + \frac{e^\varepsilon - 1}{2\delta}\right)\right)$. Let $\mathcal{B}$ be $(\alpha/2z, \beta)$-concentrated. If there exists $Y^*$ with the property that for all $i \in [t], \mathrm{dist}(Y^*, Y_i) < \alpha/2z$, then $\mathbb{P}\left[\mathrm{dist}(\widetilde{Y}, Y^*) > \alpha\right] \le \beta$.*

The utility guarantee asserts that if the outcome of all non-private procedures are close to each other, then the output of the PPE will be close to those non-private outcomes.

**Remark 2.4.** *Let $T_{\mathcal{A}}$ be the running time of the algorithm $\mathcal{A}$ in Line 2, $T_{\mathrm{dist}}$ be the time to compute $\mathrm{dist}(Y_i, Y_j)$ for any $Y_i, Y_j \in \mathcal{Y}$ in Line 3, and $T_{\mathcal{B}}$ be the time to compute $\widetilde{Y}$ in Line 9. Then Algorithm 1 runs in time $O(t \cdot T_{\mathcal{A}} + t^2 \cdot T_{\mathrm{dist}} + T_{\mathcal{B}})$. We will see that $T_{\mathcal{A}}, T_{\mathcal{B}}$, and $T_{\mathrm{dist}}$ can be polynomially bounded for GMMs.*

To apply Algorithm 1 for private learning of GMMs, we need to introduce a masking mechanism for them.

In order to do that, we start by defining how one can convert a masking mechanism for a component to one for mixtures (Section 3). We then define a masking mechanism for a single Gaussian component (presented in Section 4). Finally, we apply this to come up with a masking mechanism for GMMs as shown in Section 5.

## 3 Masking Mixtures

The goal of this section is to show how to "lift" a masking mechanism for a single component to a masking mechanism for mixtures. We can do this by adding noise to each of the components and randomly permute the output components.

Formally, let $\mathcal{F}$ denote a space and let $\mathcal{F}^k = \mathcal{F} \times \ldots \times \mathcal{F}$ ($k$ times). The following definition is useful in defining the distance between two mixtures, as it is invariant to the order of components.

**Definition 3.1.** *Let $\mathrm{dist}$ denote a distance function on $\mathcal{F}$. We define $\mathrm{dist}^k \colon \mathcal{F}^k \times \mathcal{F}^k \to \mathbb{R}_{\ge 0}$ as*

$$\mathrm{dist}^k((F_1, \ldots, F_k), (F'_1, \ldots, F'_k)) \coloneqq \min_\pi \max_{i \in [k]} \mathrm{dist}(F_i, F'_{\pi(i)}),$$

*where the minimization is taken over all permutations $\pi$.*

Note that computing $\mathrm{dist}^k$ requires computing a minimum over all permutations $\pi$. Naively, one might assume that this requires exponential time to try all permutations. However, it turns out that one can reduce the problem of computing $\mathrm{dist}^k$ to deciding whether a perfect matching exists in a weighted bipartite graph. The details of this argument can be found in Appendix E.1.

**Lemma 3.2.** *If $T_{\mathrm{dist}}$ is the running time to compute $\mathrm{dist}$ then $\mathrm{dist}^k$ can be computed in time $O(k^2 T_{\mathrm{dist}} + k^3 \log k)$.*

The following definition is useful for extending a masking mechanism for a component to a masking mechanism for a mixture. The important thing is that the components are shuffled randomly in this mechanism, making the outcome independent of the original order of the components.

**Definition 3.3.** *Suppose that $\mathcal{B}$ is a $(\gamma, \varepsilon, \delta)$-masking mechanism for $\mathcal{F}$. We define the mechanism $\mathcal{B}^k_\sigma$ as $\mathcal{B}^k_\sigma(F_1, \ldots, F_k) = (\mathcal{B}(F_{\sigma(1)}), \ldots, \mathcal{B}(F_{\sigma(k)}))$, where $\sigma$ is a uniform random permutation.*

We also note that $\mathcal{B}^k_\sigma$ can be computed with only polynomial overhead. The proof can be found in Appendix E.2.

**Lemma 3.4.** *If $T_\mathcal{B}$ is the running time of $\mathcal{B}$ then $\mathcal{B}^k_\sigma$ can be computed in time $O(k \cdot T_\mathcal{B} + k \log k)$.*

The next lemma shows that $\mathcal{B}^k_\sigma$ is indeed a masking mechanism w.r.t. $(\mathcal{F}^k, \text{dist}^k)$ and that $\mathcal{B}^k_\sigma$ is accurate provided that $\mathcal{B}$ is accurate. The proof can be found in Appendix E.3.

**Lemma 3.5.** *If $\mathcal{B}$ is an $(\alpha, \beta)$-concentrated $(\gamma, \varepsilon, \delta)$-masking mechanism for $(\mathcal{F}, \text{dist})$ then, for any $\delta' > 0$, $\mathcal{B}^k_\sigma$ is an $(\alpha, k\beta)$-concentrated $(\gamma, \varepsilon', k\delta + \delta')$-masking mechanism for $(\mathcal{F}^k, \text{dist}^k)$ where*

$$\varepsilon' = \sqrt{2k \ln(1/\delta')}\varepsilon + k\varepsilon(e^\varepsilon - 1).$$

Recall that Theorem 2.3 requires that the distance function satisfies an $r$-restricted $z$-approximate. The following lemma shows that $\text{dist}^k$ indeed does satisfy this property provided that dist does. The proof can be found in Appendix E.4.

**Lemma 3.6.** *If $\text{dist}$ satisfies an $r$-restricted $z$-approximate triangle inequality then so does $\text{dist}^k$.*

## 4 Masking a Single Gaussian Component

In this section, we develop a masking mechanism for a single Gaussian component. For that we define a new distance measure between Gaussian components.

Let $\mathcal{F}_{\text{Comp}} = \mathbb{R} \times \mathbb{R}^d \times \mathbb{R}^{d \times d}$ (corresponding to the weight $w$, mean $\mu$, and covariance matrix $\Sigma$, respectively). Define $\text{dist}_{\text{Comp}} \colon \mathcal{F}_{\text{Comp}} \times \mathcal{F}_{\text{Comp}} \to \mathbb{R}_{\geq 0}$ as

$$\text{dist}_{\text{Comp}}((w_1, \mu_1, \Sigma_1), (w_2, \mu_2, \Sigma_2)) = \max\{|w_1 - w_2|, \text{dist}_{\text{Mean}}((\mu_1, \Sigma_1), (\mu_2, \Sigma_2)), \text{dist}_{\text{Cov}}(\Sigma_1, \Sigma_2)\},$$

where

$$\text{dist}_{\text{Cov}}(\Sigma_1, \Sigma_2) = \max\{\|\Sigma_1^{1/2}\Sigma_2^{-1}\Sigma_1^{1/2} - I_d\|_F, \|\Sigma_2^{1/2}\Sigma_1^{-1}\Sigma_2^{1/2} - I_d\|_F\}$$

and

$$\text{dist}_{\text{Mean}}((\mu_1, \Sigma_1), (\mu_2, \Sigma_2)) = \max\{\|\mu_1 - \mu_2\|_{\Sigma_1}, \|\mu_1 - \mu_2\|_{\Sigma_2}\}.$$

First, we show that $\text{dist}_{\text{Comp}}$ satisfies an approximate triangle inequality; this is useful in order to use Theorem 2.3.

**Lemma 4.1.** *$\text{dist}_{\text{Comp}}$ satisfies a 1-restricted $(3/2)$-approximate triangle inequality.*

*Proof.* For any positive-definite matrix $\Sigma$, $\|\cdot\|_\Sigma$ is a metric and thus, $\text{dist}_{\text{Mean}}$ is a metric (and therefore satisfies the 1-restricted $(3/2)$-approximate triangle inequality). Next, $\text{dist}_{\text{Cov}}$ satisfies the 1-restricted $(3/2)$-approximate triangle inequality (see Lemma A.8). A straightforward calculation concludes that, as a result, $\text{dist}_{\text{Comp}}$ also satisfies a 1-restricted $(3/2)$-approximate triangle inequality. □

The following lemma gives a masking mechanism for a single Gaussian mechanism. The proof can be found Appendix F. The mechanism essentially noises the mixing weight, the mean, and the covariance matrix separately. For noising the mixing weight, one can do this using the Gaussian mechanism. Care must be taken to noise the mean and the covariance matrix. In both cases, we use the empirical covariance matrix itself to re-scale both the mean and the covariance matrix. Note that the parameters $\gamma$ (how close the inputs must be) as well as $\eta_W, \eta_{\text{Mean}}, \eta_{\text{Cov}}$ (the noise magnitude) must be set correctly to ensure privacy and accuracy. Roughly speaking, for a vector of size $d$, we should take the noise $\eta$ to about $\alpha/\sqrt{d}$ to ensure that accuracy is within an error of $\alpha$. So $\eta_W \sim \alpha$, $\eta_{\text{Mean}} \sim \alpha/\sqrt{d}$ and $\eta_{\text{Cov}} \sim \alpha/d$. To ensure $(\varepsilon, \delta)$-DP, we need to ensure that the outputs are guaranteed to be close enough (otherwise the noise is not enough to make the inputs indistinguishable). The bottleneck comes from noising the covariance matrix which requires that $\gamma \sim \varepsilon \eta_{\text{Cov}}/\sqrt{d} \sim \alpha \varepsilon / d^{3/2}$. More details about the parameter choices can be found in Appendix F.

**Lemma 4.2.** *For $\gamma \leq \frac{\varepsilon\alpha}{C_2\sqrt{d^2(d+\ln(4/\beta))\cdot\ln(2/\delta)}}$, there exists a $(\gamma, 3\varepsilon, 3\delta)$-masking mechanism, $\mathcal{B}_{\text{COMP}}$, for $(\mathcal{F}_{\text{COMP}}, \text{dist}_{\text{COMP}})$ that is $(\alpha, 3\beta)$-concentrated, where $C_2$ is a universal constant.*

## 5 A Masking Mechanism for GMMs

In this section, we show how to mask a mixture of $k$ Gaussians. Let $\mathcal{F}_{\text{GMM}} = \mathcal{F}_{\text{COMP}} \times \ldots \times \mathcal{F}_{\text{COMP}}$ ($k$ times). Note we drop $k$ from $\mathcal{F}_{\text{GMM}}$ (and related notation below) since $k$ is fixed and implied from context. Let $\text{dist}_{\text{COMP}}$ be as defined in Eq. (4) and define the distance

$$\text{dist}_{\text{PARAM}}(\{(w_i, \mu_i, \Sigma_i)\}_{i\in[k]}, \{(w'_i, \mu'_i, \Sigma'_i)\}_{i\in[k]}) = \min_\pi \max_{i\in[k]} \text{dist}_{\text{COMP}}((w_{\pi(i)}, \mu_{\pi(i)}, \Sigma_{\pi(i)}), (w'_i, \mu'_i, \Sigma'_i)),$$

where $\pi$ is chosen from the set of all permutations over $[k]$. Now define the masking mechanism

$$\mathcal{B}_{\text{GMM}}(\{(w_i, \mu_i, \Sigma_i)\}_{i\in[k]}) = \{\mathcal{B}_{\text{COMP}}(w_{\sigma(i)}, \mu_{\sigma(i)}, \Sigma_{\sigma(i)})\}_{i\in[k]},$$

where $\mathcal{B}_{\text{COMP}}$ is the masking mechanism from Lemma 4.2 and $\sigma$ is a permutation chosen uniformly at random from the set of all permutations over $[k]$. In words, $\mathcal{B}_{\text{GMM}}$ applies the masking mechanism $\mathcal{B}_{\text{COMP}}$ from Section 4 to each component separately and then permutes the components. To summarize the entire masking mechanism for GMMs, we provide pseudocode in Algorithm 2.

The following lemma asserts that $\mathcal{B}_{\text{GMM}}$ is indeed a masking mechanism. At a high-level, it follows by combining Lemma 4.2 with Lemma 3.5. The details can be found in Appendix G.1.

**Lemma 5.1.** *Let $\varepsilon < \ln(2)/3$. There is a sufficiently large constant $C_2$ such that for $\gamma \leq \frac{\varepsilon\alpha}{C_2\sqrt{k\ln(2/\delta)}\sqrt{d^2(d+\ln(12k/\beta))\cdot\ln(12k/\delta)}}$, $\mathcal{B}_{\text{GMM}}$ is a $(\gamma, \varepsilon, \delta)$-masking mechanism with respect to $(\mathcal{F}_{\text{GMM}}, \text{dist}_{\text{PARAM}})$. Moreover, $\mathcal{B}_{\text{GMM}}$ is $(\alpha, \beta)$-concentrated.*

Note that $\text{dist}_{\text{PARAM}}$ also satisfies a 1-restricted $(3/2)$-approximate triangle inequality since $\text{dist}_{\text{COMP}}$ does (see Appendix G.2 for a proof).

**Lemma 5.2.** *$\text{dist}_{\text{PARAM}}$ satisfies a 1-restricted $(3/2)$-approximate triangle inequality.*

## 6 Privately Learning GMMs

At this point, we have everything we need to develop a private algorithm for learning the parameters of a GMM. First, we define the problem more formally.

**Definition 6.1** (PAC Learning of Parameters of GMMs). *Let $\mathcal{F} = \left\{\left(w_i^j, \mu_i^j, \Sigma_i^j\right)_{i=1}^k\right\}^j$ be a class of $d$-dimensional GMMs with $k$ components[5]. Let $\mathcal{A}$ be function that receives a sequence $S$ of instances in $\mathbb{R}^d$ and outputs a mixture $\hat{F} = (\hat{w}_i, \hat{\mu}_i, \hat{\Sigma}_i)_{i=1}^k$. Let $m: (0,1)^2 \times \mathbb{N}^2 \to \mathbb{N}$. We say $\mathcal{A}$ learns the parameters of $\mathcal{F}$ with $m$ samples if for every $\alpha, \beta \in (0,1)$ and every $F \in \mathcal{F}$, if $S$ is an i.i.d. sample of size $m(\alpha, \beta, k, d)$ from $F$, then $\text{dist}_{\text{GMM}}(F, \hat{F}) < \alpha$ with probability at least $1 - \beta$.*

Plugging the masking mechanism developed in Section 5 (in particular, Lemma 5.1 and Lemma 5.2) into PPE (Theorem 2.3) gives a private to non-private reduction for GMMs.

---
[5]For examples, it is standard to pick $\mathcal{F}$ to be those GMMs that are separable/identifiable.

**Algorithm 2** GMM Masking Mechanism
---
**Input:** GMM given by $\{(w_i, \mu_i, \Sigma_i)\}_{i\in[k]}$ and parameters $\eta_{\text{W}}, \eta_{\text{Mean}}, \eta_{\text{Cov}} > 0$

1: **function** $\mathcal{R}_{\text{W}}(w)$  ▷ Noise mixing weights
2:     **Return** $\max(0, w + \eta_{\text{W}} g)$ where $g \sim \mathcal{N}(0, 1)$.
3: **end function**
4: **function** $\mathcal{R}_{\text{Mean}}(\mu, \Sigma)$  ▷ Noise mean
5:     **Return** $\mu + \eta_{\text{Mean}} g$ where $g \sim \mathcal{N}(0, \Sigma)$
6: **end function**
7: **function** $\mathcal{R}_{\text{Cov}}(\Sigma)$  ▷ Noise covariance
8:     Let $G \in \mathbb{R}^{d \times d}$ matrix with independent $\mathcal{N}(0, 1)$ entries.
9:     **Return** $\Sigma^{1/2}(I_d + \eta_{\text{Cov}} G)(I_d + \eta_{\text{Cov}} G)^\top \Sigma^{1/2}$
10: **end function**
11: **function** $\mathcal{B}_{\text{Comp}}(w, \mu, \Sigma)$  ▷ Mask component
12:     **Return** $(\mathcal{R}_{\text{W}}(w), \mathcal{R}_{\text{Mean}}(\mu, \Sigma), \mathcal{R}_{\text{Cov}}(\Sigma))$
13: **end function**
14: **function** $\mathcal{B}_{\text{Gmm}}(\{(w_i, \mu_i, \Sigma_i)\}_{i \in [k]})$  ▷ Mask GMM
15:     Let $\sigma$ be uniformly random permutation.
16:     $\{(\hat{w}_i, \hat{\mu}, \hat{\Sigma}_i)\} \leftarrow \{\mathcal{B}_{\text{Comp}}(w_{\sigma(i)}, \mu_{\sigma(i)}, \Sigma_{\sigma(i)})\}$.
17:     Normalize: $\hat{w}_i \leftarrow \hat{w}_i / \sum_{i \in [k]} \hat{w}_i$.
18:     **Return** $\{(\hat{w}_i, \hat{\mu}, \hat{\Sigma}_i)\}_{i \in [k]}$.
19: **end function**

**Theorem 6.2** (Private to Non-Private Reduction). *Let $\mathcal{F}$ be a subclass of GMMs with $k$ components in $\mathbb{R}^d$. Let $\mathcal{A}$ be a non-private Algorithm that PAC learns the parameters of $\mathcal{F}$ with respect to* $\text{dist}_{GMM}$ *using $m_{\text{non-private}}(\alpha, \beta, k, d)$ samples. Then for every $\varepsilon < \ln(2)/3$, $\delta \in (0, 1)$, $\gamma \leq \frac{\varepsilon \alpha}{C_2 \sqrt{k \ln(2/\delta)} \sqrt{d^2(d + \ln(12k/\beta)) \cdot \ln(12k/\delta)}}$ for a sufficiently large constant $C$ and $t = \max\{5, \lceil \frac{20}{\varepsilon} \ln(1 + \frac{e^\varepsilon - 1}{2\delta}) \rceil\}$, there is a learner $\mathcal{A}_{\text{private}}$ with the following properties:*

1. *$\mathcal{A}_{\text{private}}$ is $(2\varepsilon, 4e^\varepsilon \delta)$-DP.*
2. *$\mathcal{A}_{\text{private}}$ PAC learns the parameters of $\mathcal{F}$ using $O(m_{\text{non-private}}(\gamma, \beta/2t, k, d) \log(1/\delta)/\varepsilon)$ samples.*
3. *$\mathcal{A}_{\text{private}}$ runs in time $O((\log(1/\delta)/\varepsilon) \cdot T_\mathcal{A} + (\log(1/\delta)/\varepsilon)^2 \cdot (k^2 d^3 + k^3 \log k))$, where $T_\mathcal{A}$ is the running time for the non-private algorithm.*

To prove Theorem 6.2, we require the following lemma whose proof can be found in Appendix H.

**Lemma 6.3.** *Let $F = (w_i, \mu_i, \Sigma_i)_{i=1}^k$ and $F' = (w'_i, \mu'_i, \Sigma'_i)_{i=1}^k$ be two d-dimensional GMMs where $\Sigma_i$ and $\Sigma'_i$ are positive-definite matrices. Suppose that $\text{dist}_{GMM}(F, F') < \frac{1}{600}$. Then $\frac{1}{200} \text{dist}_{PARAM}(F, F') \leq \text{dist}_{GMM}(F, F') \leq \frac{1}{\sqrt{2}} \text{dist}_{PARAM}(F, F')$.*

*Proof of Theorem 6.2.* Let $z = 3/2$, $r = 1$, and $t \geq \frac{20}{\varepsilon} \ln\left(1 + \frac{e^\varepsilon - 1}{2\delta}\right) = O(\log(1/\delta)/\varepsilon)$. We run Algorithm 1 with the following.

- For the non-private algorithm $\mathcal{A}$, we use the algorithm from Theorem 6.5 with accuracy parameter $\alpha/2z$ and failure probability $\beta/2t$.
- For the masking mechanism, we use the $(r, \varepsilon, \delta)$-masking mechanism $\mathcal{B}_{\text{Gmm}}$ which is defined in Lemma 5.1. Further, this mechanism is $(\alpha/2z, \beta/2)$-concentrated.
- Finally, note that the distance function $\text{dist}_{\text{PARAM}}$ satisfies the $z$-approximate $r$-restricted triangle inequality (Lemma 5.2).

Let $F^*$ be the true GMM. Let $F_i$ be the estimated GMMs computed by $\mathcal{A}$ in Line 2 of Algorithm 1. Then the first item above guarantees that $\text{dist}_{\text{PARAM}}(F^*, F_i) \leq \alpha/2z$ for all $i \in [t]$ with probability at least $1 - \beta/2$.

We thus conclude that we have a private algorithm for learning GMMs that is $(2\varepsilon, 4e^\varepsilon \delta)$-DP and that returns $\widetilde{F}$ satisfying $\text{dist}_{\text{PARAM}}(\widetilde{F}, F^*) \leq \alpha$ with probability $1 - \beta$. By Lemma 6.3, we further conclude that $\text{dist}_{\text{GMM}}(\widetilde{F}, F^*) \leq O(\alpha)$ with probability $1 - \beta$.

It remains to check the sample complexity and computational complexity of our algorithm. Since we run $t$ independent instances of the non-private algorithm $\mathcal{A}$, we require $t \cdot m_{\text{PRIVATE}}(\alpha/2z, \beta/2t, k, d) = O(m_{\text{PRIVATE}}(\alpha/2z, \beta/2t, k, d) \cdot \log(1/\delta)/\varepsilon)$ samples. Finally, we bound the running time. Lemma 3.4 shows that the running time to apply the masking mechanism is $O(k \cdot d^3 + k \log k)$ and Lemma 3.2 shows that the running time to compute dist is $O(k^2 d^3 + k^3 \log k)$. The claimed running time now follows from Remark 2.4. $\square$

## 6.1 Application

As a concrete application, we apply Theorem 6.2 with the algorithm of Moitra & Valiant [MV10] to obtain the first private algorithm for learning the parameters of a GMM with sample and computational complexity that is polynomial in $d$ (for a fixed $k$) with minimal separation assumptions. Note that our algorithm does not require any boundedness assumptions on the parameters.

**Definition 6.4** ($\gamma$-Statistically Learnable [MV10]). We say a GMM $F = (w_i, \mu_i, \Sigma_i)_{i=1}^k$ is $\gamma$-statistically learnable if (i) $\min_i w_i \geq \gamma$ and (ii) $\min_{i \neq j} d_{\text{TV}}(\mathcal{N}(\mu_i, \Sigma_i), \mathcal{N}(\mu_j, \Sigma_j)) \geq \gamma$.

If a GMM is $\gamma$-statistically learnable, we will be able to recover its components accurately.

**Theorem 6.5** (Non-private Learning of GMMs [MV10]). *There exists an algorithm $\mathcal{A}$ and a function $m_\mathcal{A}(d, k, \alpha, \beta)$ with the following guarantee. Fix $\alpha, \beta \in (0, 1)$, $k, d \in \mathbb{N}$.*

- *For fixed $k$, the sample complexity $m_\mathcal{A}(d, k, \alpha, \beta)$ is polynomial in $d/\alpha\beta$.*
- *For fixed $k$, $\mathcal{A}$ runs in time $\text{poly}(d/\alpha\beta)$.*
- *Let $\mathcal{F}^*$ be an $\alpha$-statistically learnable subclass of GMMs with $k$ components in $\mathbb{R}^d$ and let $F^* \in \mathcal{F}^*$. Given an i.i.d. sample $D$ of size $m_\mathcal{A}(d, k, \alpha, \beta)$ drawn from $F^*$, with probability at least $1 - \beta$, $\mathcal{A}$ return $\hat{F}$ such that $\text{dist}_{GMM}(\hat{F}, F^*) \leq \alpha$.*

The following corollary follows immediately by plugging Theorem 6.5 into Theorem 6.2.

**Corollary 6.6.** *There exists an algorithm $\mathcal{A}$ and a function $m_\mathcal{A}(d, k, \alpha, \beta, \varepsilon, \delta)$ with the following guarantee. Fix $\alpha, \beta, \varepsilon, \delta \in (0, 1)$, $k, d \in \mathbb{N}$.*

- *$\mathcal{A}$ is $(\varepsilon, \delta)$-DP.*
- *For fixed $k$, the sample complexity $m_\mathcal{A}(d, k, \alpha, \beta, \varepsilon, \delta)$ is polynomial in $d \log(1/\delta)/\alpha\beta\varepsilon$.*
- *For fixed $k$, $\mathcal{A}$ runs in time $\text{poly}(d \log(1/\delta)/\alpha\beta\varepsilon)$.*
- *Let $\mathcal{F}^*$ be an $\alpha$-statistically learnable subclass of GMMs with $k$ components in $\mathbb{R}^d$ and let $F^* \in \mathcal{F}^*$. Given an i.i.d. sample $D$ of size $m_\mathcal{A}(d, k, \alpha, \beta, \varepsilon, \delta)$ drawn from $F^*$, with probability at least $1 - \beta$, $\mathcal{A}$ return $\hat{F}$ such that $\text{dist}_{GMM}(\hat{F}, F^*) \leq \alpha$.*

**Acknowledgements.** The authors would like to thank Abbas Mehrabian for pointing out a mistake in an earlier version Theorem 1.8. The authors would also like to the anonymous reviewers for many insightful comments. Hassan Ashtiani was supported by an NSERC Discovery grant.

## A  Standard Facts

**Fact A.1.** *Let $X_1, X_2, Y_1, Y_2$ be random variables such that $X_1, X_2$ (resp. $Y_1, Y_2$) are independent. Then $d_{\mathrm{TV}}\left((X_1, X_2), (Y_1, Y_2)\right) \leq d_{\mathrm{TV}}\left(X_1, Y_1\right) + d_{\mathrm{TV}}\left(X_2, Y_2\right)$.*

**Fact A.2.** *Let $X, Y$ be random variables. For any measurable function $f$, $d_{\mathrm{TV}}\left(f(X), f(Y)\right) \leq d_{\mathrm{TV}}\left(X, Y\right)$.*

The equality in the following fact is standard; for example, see Equation 2.3 in Williams & Rasmussen [WR06]. For the inequality, see the proof of Lemma 2.9 in Ashtiani et al. [ABHLMP20].

**Fact A.3.** *Let $\mu_1, \mu_2 \in \mathbb{R}^d$ and $\Sigma_1, \Sigma_2 \succ 0$. Then*

$$D_{\mathrm{KL}}\left(\mathcal{N}(\mu_1, \Sigma_1) \parallel \mathcal{N}(\mu_2, \Sigma_2)\right) = \frac{1}{2}\left[\operatorname{tr}(\Sigma_2^{-1}\Sigma_1 - I) + (\mu_2 - \mu_1)^\top \Sigma_2^{-1}(\mu_2 - \mu_1) - \ln\det(\Sigma_2^{-1}\Sigma_1)\right].$$

*Moreover, suppose that all the eigenvalues of $\Sigma_2^{-1/2}\Sigma_1\Sigma_2^{-1/2}$ are at least $\frac{1}{2}$. Then*

$$D_{\mathrm{KL}}\left(\mathcal{N}(\mu_1, \Sigma_1) \parallel \mathcal{N}(\mu_2, \Sigma_2)\right) \leq \frac{1}{2}\left[\|\Sigma_2^{-1/2}\Sigma_1\Sigma_2^{-1/2} - I\|_F^2 + (\mu_2 - \mu_1)^\top \Sigma_2^{-1}(\mu_2 - \mu_1)\right].$$

**Lemma A.4** (Pinsker's Inequality). *Let $P$ and $Q$ be two distributions for which KL-divergence is defined. Then $d_{\mathrm{TV}}(P,Q) \leq \sqrt{0.5 D_{\mathrm{KL}}(P \parallel Q)}$.*

**Lemma A.5** ([LM00], Lemma 1). *Let $g_1, \ldots, g_k$ be i.i.d. $\mathcal{N}(0,1)$ random variables. Then*

$$\mathbb{P}\left[\sum_{i=1}^{k} g_i^2 \geq k + 2\sqrt{kt} + 2t\right] \leq e^{-t}.$$

**Lemma A.6** ([AL22], Lemma D.2). *Let $\mu_1, \mu_2 \in \mathbb{R}^d$ and let $\Sigma_1, \Sigma_2$ be full-rank $d \times d$ PSD matrices. Let $Y \sim \mathcal{N}(\mu_1, \Sigma_1)$. Then*

$$\begin{aligned}\mathcal{L}_{\mathcal{N}(\mu_1,\Sigma_1)\|\mathcal{N}(\mu_2,\Sigma_2)}(Y) &\leq D_{\mathrm{KL}}\left(\mathcal{N}(\mu_1,\Sigma_1) \parallel \mathcal{N}(\mu_2,\Sigma_2)\right) \\ &+ 2\|\Sigma_1^{1/2}\Sigma_2^{-1}\Sigma_1^{1/2} - I_d\|_F \cdot \sqrt{\ln(2/\delta)} + 2\|\Sigma_1^{1/2}\Sigma_2^{-1}\Sigma_1^{1/2} - I_d\| \cdot \ln(2/\delta) \\ &+ \|\Sigma_1^{1/2}\Sigma_2^{-1}\Sigma_1^{1/2}\| \cdot \|\Sigma_1^{-1/2} \cdot (\mu_2 - \mu_1)\|_2 \cdot \sqrt{2\ln(2/\delta)}\end{aligned} \quad (A.1)$$

*with probability at least $1 - \delta$.*

**Fact A.7.** *For $x \in (0, \ln(2))$, we have $e^x \leq 1 + 2x$*

*Proof.* Consider the function $f(x) = 1 + 2x - e^x$. Then $f''(x) = -e^x$ so $f$ is concave. Note that $f(0) = 0$ and $f(\ln(2)) = 1 + 2\ln(2) - 2 > 0$ so $f(x) \geq 0$ for $x \in [0, \ln(2)]$ (by concavity). □

**Lemma A.8** ([AL22], Lemma 3.2). *Let $\mathcal{S}^d$ be the set of all $d \times d$ positive definite matrices. For $A, B \in \mathcal{S}^d$ let $\mathrm{dist}(A,B) = \max\{\|A^{-1/2}BA^{-1/2} - I\|, \|B^{-1/2}AB^{-1/2} - I\|\}$. Then $(\mathcal{S}^d, \mathrm{dist})$ is a semimetric space which satisfies a $(3/2)$-approximate $1$-restricted triangle inequality.*

**Fact A.9.** *Let $\Sigma_1, \Sigma_2$ be $d \times d$ positive-definite matrices. Then $\Sigma_2^{-1/2}\Sigma_1\Sigma_2^{-1/2}$ and $\Sigma_1^{1/2}\Sigma_2^{-1}\Sigma_1^{1/2}$ have the same spectrum.*

*Proof.* Suppose that $x \in \mathbb{R}^d$ is an eigenvector of $\Sigma_2^{-1/2}\Sigma_1\Sigma_2^{-1/2}$ with eigenvalue $\lambda$. Let $y = \Sigma_1^{1/2}\Sigma_2^{-1/2}x$. Then

$$\Sigma_1^{1/2}\Sigma_2^{-1}\Sigma_1^{1/2}y = \Sigma_1^{1/2}\Sigma_2^{-1/2}\left(\Sigma_2^{-1/2}\Sigma_1\Sigma_2^{-1/2}\right)x = \lambda \Sigma_1^{1/2}\Sigma_2^{-1/2}x = \lambda y.$$

In other words, $y$ is an eigenvector of $\Sigma_1^{1/2}\Sigma_2^{-1}\Sigma_1^{1/2}$ with eigenvalue $\lambda$. □

# B  TV Distance of Gaussian Distributions

In this section, we prove the following theorem.

**Theorem 1.8.** *Let $\mu_1, \mu_2 \in \mathbb{R}^d$ and $\Sigma_1, \Sigma_2$ be $d \times d$ positive-definite matrices. Suppose that we have $d_{\mathrm{TV}}(\mathcal{N}(\mu_1, \Sigma_1), \mathcal{N}(\mu_2, \Sigma_2)) < \frac{1}{600}$. Let*

$$\Delta = \max\left\{\|\Sigma_1^{-1/2}\Sigma_2\Sigma_1^{-1/2} - I_d\|_F, \|\Sigma_1^{-1/2}(\mu_1 - \mu_2)\|_2\right\}.$$

*Then*

$$\frac{1}{200}\Delta \leq d_{\mathrm{TV}}\left(\mathcal{N}(\mu_1, \Sigma_1), \mathcal{N}(\mu_2, \Sigma_2)\right) \leq \frac{1}{\sqrt{2}}\Delta.$$

Our proof makes use of the following two theorems from [DMR18].

**Theorem B.1** ([DMR18], Theorem 1.1)**.** *Let $\mu \in \mathbb{R}^d$, $\Sigma_1, \Sigma_2 \in \mathcal{S}^d$, The total variation distance between Gaussians with the same mean is bounded by*

$$\frac{\min\left\{1, \|\Sigma_1^{-1/2}\Sigma_2\Sigma_1^{-1/2} - I_d\|_F\right\}}{100} \leq d_{\mathrm{TV}}\left(\mathcal{N}(\mu, \Sigma_1), \mathcal{N}(\mu, \Sigma_2)\right)$$

**Theorem B.2** ([DMR18], Theorem 1.3)**.** *The total variation distance between one-dimensional Gaussians is bounded by*

$$\frac{1}{200}\min\left\{1, \max\left\{\frac{|\sigma_1^2 - \sigma_2^2|}{\sigma_1^2}, \frac{40|\mu_1 - \mu_2|}{\sigma_1}\right\}\right\} \leq d_{\mathrm{TV}}\left(\mathcal{N}(\mu_1, \sigma_1^2), \mathcal{N}(\mu_2, \sigma_2^2)\right) \leq \frac{3|\sigma_1^2 - \sigma_2^2|}{2\sigma_1^2} + \frac{|\mu_1 - \mu_2|}{2\sigma_1}.$$

**Lemma B.3.** $d_{\mathrm{TV}}\left(\mathcal{N}(0, \Sigma_1), \mathcal{N}(0, \Sigma_2)\right) \leq 2 \cdot d_{\mathrm{TV}}\left(\mathcal{N}(\mu_1, \Sigma_1), \mathcal{N}(\mu_2, \Sigma_2)\right).$

*Proof.* Let $X_1, X_2, Y_1, Y_2$ be independent random variables where $X_1, X_2 \sim \mathcal{N}(\mu_1, \Sigma_1)$ and $Y_1, Y_2 \sim \mathcal{N}(\mu_2, \Sigma_2)$. Applying Fact A.1 gives

$$\begin{aligned}
d_{\mathrm{TV}}\left((X_1, X_2), (Y_1, Y_2)\right) &\leq d_{\mathrm{TV}}(X_1, Y_1) + d_{\mathrm{TV}}(X_2, Y_2) \\
&= d_{\mathrm{TV}}\left(\mathcal{N}(\mu_1, \Sigma_1), \mathcal{N}(\mu_2, \Sigma_2)\right) + d_{\mathrm{TV}}\left(\mathcal{N}(\mu_1, \Sigma_1), \mathcal{N}(\mu_2, \Sigma_2)\right) \\
&= 2 \cdot d_{\mathrm{TV}}\left(\mathcal{N}(\mu_1, \Sigma_1), \mathcal{N}(\mu_2, \Sigma_2)\right).
\end{aligned}$$

Now, let $X = (X_1, X_2)$ and $Y = (Y_1, Y_2)$, and define the function $f(X) = f(X_1, X_2) = (X_1 - X_2)/\sqrt{2}$. Then by applying Fact A.2 we have

$$d_{\mathrm{TV}}\left(f(X_1, X_2), f(Y_1, Y_2)\right) \leq d_{\mathrm{TV}}\left((X_1, X_2), (Y_1, Y_2)\right) \leq 2 \cdot d_{\mathrm{TV}}\left(\mathcal{N}(\mu_1, \Sigma_1), \mathcal{N}(\mu_2, \Sigma_2)\right).$$

Note that if $X_1, X_2 \sim \mathcal{N}(\mu_1, \Sigma_1)$ then $f(X) \sim \mathcal{N}(0, \Sigma_1)$. Therefore we have

$$d_{\mathrm{TV}}\left(\mathcal{N}(0, \Sigma_1), \mathcal{N}(0, \Sigma_2)\right) \leq 2 \cdot d_{\mathrm{TV}}\left(\mathcal{N}(\mu_1, \Sigma_1), \mathcal{N}(\mu_2, \Sigma_2)\right),$$

as required. □

**Lemma B.4.** *Let $\mu \in \mathbb{R}^d$. If $d_{\mathrm{TV}}\left(\mathcal{N}(0, I_d), \mathcal{N}(\mu, I_d)\right) \leq 3\alpha < 1/200$ then $\|\mu\|_2 \leq 15\alpha$.*

*Proof.* Let $g_1 \sim \mathcal{N}(0, I_d)$, $g_2 \sim \mathcal{N}(\mu, I_d)$ and $v = \mu/\|\mu\|_2$. Note that $v^\top g_1 \sim \mathcal{N}(0, 1)$ and $v^\top g_2 \sim \mathcal{N}(\|\mu\|_2, 1)$. Applying Fact A.2 (with $f(x) = v^\top x$) we have

$$d_{\mathrm{TV}}\left(\mathcal{N}(0, 1), \mathcal{N}(\|\mu\|_2, 1)\right) \leq d_{\mathrm{TV}}\left(\mathcal{N}(0, I_d), \mathcal{N}(\mu, I_d)\right) \leq 3\alpha < 1/200.$$

Applying Theorem B.2 on the left side, we have

$$\frac{1}{200}\min\left\{1, 40\|\mu\|_2\right\} \leq d_{\mathrm{TV}}\left(\mathcal{N}(0, 1), \mathcal{N}(\|\mu\|_2, 1)\right) \leq d_{\mathrm{TV}}\left(\mathcal{N}(0, I_d), \mathcal{N}(\mu, I_d)\right) \leq 3\alpha < 1/200.$$

Note that this implies $\min\{1, 40\|\mu\|_2\} = 40\|\mu\|_2 < 1$. Therefore we conclude that $\|\mu\|_2 \leq 15\alpha$. □

**Lemma B.5.** *Let $\mu_1, \mu_2 \in \mathbb{R}^d$ and $\Sigma_1, \Sigma_2$ be full-rank $d \times d$ positive-definite matrices. Suppose that*

$$d_{\mathrm{TV}}\left(\mathcal{N}(\mu_1, \Sigma_1), \mathcal{N}(\mu_2, \Sigma_2)\right) \leq \alpha < \frac{1}{600}.$$

*Then (i) $\|\Sigma_1^{-1/2}\Sigma_2\Sigma_1^{-1/2} - I_d\|_F \leq 200\alpha$ and (ii) $\|\Sigma_1^{-1/2}(\mu_1 - \mu_2)\|_2 \leq 15\alpha$.*

*Proof.* (i) Starting from the assumption

$$d_{\text{TV}}\left(\mathcal{N}(\mu_1, \Sigma_1), \mathcal{N}(\mu_2, \Sigma_2)\right) \leq \alpha < \frac{1}{600},$$

we apply Lemma B.3 to obtain

$$d_{\text{TV}}\left(\mathcal{N}(0, \Sigma_1), \mathcal{N}(0, \Sigma_2)\right) \leq 2\alpha < \frac{1}{300}.$$

Applying Theorem B.1 gives

$$\min\left\{1, \|\Sigma_1^{-1/2} \Sigma_2 \Sigma_1^{-1/2} - I_d\|_F\right\} \leq 100 \cdot d_{\text{TV}}\left(\mathcal{N}(0, \Sigma_1), \mathcal{N}(0, \Sigma_2)\right) \leq 200\alpha < \frac{1}{3}.$$

Note that the inequality implies that $\min\left\{1, \|\Sigma_1^{-1/2} \Sigma_2 \Sigma_1^{-1/2} - I_d\|_F\right\} = \|\Sigma_1^{-1/2} \Sigma_2 \Sigma_1^{-1/2} - I_d\|_F$. We conclude that

$$\|\Sigma_1^{-1/2} \Sigma_2 \Sigma_1^{-1/2} - I_d\|_F \leq 200\alpha.$$

This proves the first assertion.

(ii) By the triangle inequality, we have

$$d_{\text{TV}}\left(\mathcal{N}(\mu_1, \Sigma_1), \mathcal{N}(\mu_2, \Sigma_1)\right) \leq d_{\text{TV}}\left(\mathcal{N}(\mu_1, \Sigma_1), \mathcal{N}(\mu_2, \Sigma_2)\right) + d_{\text{TV}}\left(\mathcal{N}(\mu_2, \Sigma_1), \mathcal{N}(\mu_2, \Sigma_2)\right)$$
$$= d_{\text{TV}}\left(\mathcal{N}(\mu_1, \Sigma_1), \mathcal{N}(\mu_2, \Sigma_2)\right) + d_{\text{TV}}\left(\mathcal{N}(0, \Sigma_1), \mathcal{N}(0, \Sigma_2)\right).$$

Our hypothesis is that

$$d_{\text{TV}}\left(\mathcal{N}(\mu_1, \Sigma_1), \mathcal{N}(\mu_2, \Sigma_2)\right) \leq \alpha < \frac{1}{600},$$

which, by Lemma B.3, implies that

$$d_{\text{TV}}\left(\mathcal{N}(0, \Sigma_1), \mathcal{N}(0, \Sigma_2)\right) \leq 2\alpha < \frac{1}{300}.$$

Thus, we have

$$d_{\text{TV}}\left(\mathcal{N}(\mu_1, \Sigma_1), \mathcal{N}(\mu_2, \Sigma_1)\right) \leq 3\alpha < \frac{1}{200}.$$

Furthermore, since bijective mappings preserve the total variation distance, we have

$$d_{\text{TV}}\left(\mathcal{N}(\mu_1, \Sigma_1), \mathcal{N}(\mu_2, \Sigma_1)\right) = d_{\text{TV}}\left(\mathcal{N}(\Sigma_1^{-1/2}\mu_1, I_d), \mathcal{N}(\Sigma_1^{-1/2}\mu_2, I_d)\right)$$
$$= d_{\text{TV}}\left(\mathcal{N}(0, I_d), \mathcal{N}(\Sigma_1^{-1/2}(\mu_1 - \mu_2), I_d)\right).$$

Finally, applying Lemma B.4 gives $\|\Sigma_1^{-1/2}(\mu_1 - \mu_2)\|_2 \leq 15\alpha$. □

*Proof of Theorem 1.8.* The lower bound follows from Lemma B.5.

The upper bound is a standard application of Pinsker's Inequality but we provide the proof for completeness. By Lemma B.5(i) the eigenvalues of $\Sigma_2^{-1/2} \Sigma_1 \Sigma_2^{-1/2}$ are strictly larger than $1/2$ (note we swapped the indices in the application of Lemma B.5). Therefore, using Fact A.3 we know that

$$D_{\text{KL}}\left(\mathcal{N}(\mu_1, \Sigma_1) \| \mathcal{N}(\mu_2, \Sigma_2)\right) \leq \frac{1}{2}\left[\|\Sigma_2^{-1/2} \Sigma_1 \Sigma_2^{-1/2} - I\|_F^2 + (\mu_2 - \mu_1)^\top \Sigma_2^{-1} (\mu_2 - \mu_1)\right].$$

Using Pinsker's inequality (Lemma A.4) and the known fact $\|\mu\|_\Sigma = \|\Sigma^{-1/2}\mu\|_2 = \sqrt{(\mu^T \Sigma^{-1} \mu)}$ we have

$$d_{\text{TV}}\left(\mathcal{N}(\mu_1, \Sigma_1), \mathcal{N}(\mu_2, \Sigma_2)\right) \leq \frac{1}{2}\sqrt{\left[\|\Sigma_2^{-1/2} \Sigma_1 \Sigma_2^{-1/2} - I\|_F^2 + (\mu_2 - \mu_1)^\top \Sigma_2^{-1}(\mu_2 - \mu_1)\right]} \leq \frac{\Delta}{\sqrt{2}}$$

which concludes the proof. □

# C  Standard Facts about Differential Privacy

**Definition C.1.** Let $\mathcal{D}_1, \mathcal{D}_2$ be two continuous distributions defined on $\mathbb{R}^d$ and let $f_1, f_2$ be the respective density functions. We use $\mathcal{L}_{\mathcal{D}_1 \| \mathcal{D}_2} \colon \mathbb{R}^d \to \mathbb{R}$ to denote the logarithm of the likelihood ratio, i.e. for any $x \in \mathbb{R}^d$,

$$\mathcal{L}_{\mathcal{D}_1 \| \mathcal{D}_2}(x) := \ln \frac{f_1(x)}{f_2(x)}. \tag{C.1}$$

Below definition has $D, D'$ which are different in single individual data and function f can capture the change in magnitude at the worst case.

**Definition C.2** ($L_1$-Sensitivity ([DR$^+$14], Definition 3.1))**.** The $L_1$-sensitivity of a function $f \colon \mathcal{X}^n \to \mathbb{R}^k$ is defined as:

$$\Delta(f) = \max_{D, D' \in \mathcal{X}^n \,:\, d_H(D,D') \leq 1} ||f(D) - f(D')||_1$$

where $d_H$ is Hamming distance identified in Definition 1.5

**Definition C.3** ($L_2$-Sensitivity ([DR$^+$14], Definition 3.8))**.** The $L_2$-sensitivity of a function $f \colon \mathcal{X}^n \to \mathbb{R}^k$ is defined as:

$$\Delta_2(f) = \max_{D, D' \in \mathcal{X}^n \,:\, d_H(D,D') \leq 1} ||f(D) - f(D')||_2$$

where $d_H$ is Hamming distance identified in Definition 1.5

The Gaussian Mechanism with parameter $\sigma$ adds noise scaled to $\mathcal{N}(0, \sigma^2)$ to each of the $d$ components of the output.

**Theorem C.4** (Gaussian Mechanism ([DR$^+$14], Theorem 3.22))**.** *Let $\varepsilon \in (0,1)$ be arbitrary. For $c^2 > 2\ln(1.25/\delta)$, the Gaussian Mechanism with parameter $\sigma \geq c\Delta_2 f/\varepsilon$ is $(\varepsilon, \delta)$-differentially private.*

The amount of noise necessary to ensure differential privacy for a given function depends on the sensitivity of the function. In other words, we can guarantee privacy using additive noise if the sensitivity of the function is bounded. The sensitivity of a function reflects the amount the function's output will change when its input changes.

**Definition C.5** (Truncated Laplace distribution)**.** It is denoted by $\mathrm{TLap}(\Delta, \varepsilon, \delta)$ whose probability density function is given by

$$f_{\mathrm{TLap}(\Delta, \varepsilon, \delta)}(x) := \begin{cases} Be^{-|x|/\lambda} & x \in [-A, A] \\ 0 & x \notin [-A, A] \end{cases},$$

where $\lambda = \frac{\Delta}{\varepsilon}$, $A = \frac{\Delta}{\varepsilon} \ln\left(1 + \frac{e^\varepsilon - 1}{2\delta}\right)$, $B = \frac{1}{2\lambda(1 - e^{-A/\lambda})}$.

**Theorem C.6** ([GDGK18], Theorem 1)**.** *Suppose that $q \colon \mathcal{X}^n \to \mathbb{R}$ is a function with $L_1$-sensitivity $\Delta$. Then the mechanism $q(x) + Y$ where $Y \sim \mathrm{TLap}(\Delta, \varepsilon, \delta)$ is $(\varepsilon, \delta)$-DP.*

**Theorem C.7** (Advanced Composition [DRV10])**.** *Let $\mathcal{D}_1, \ldots, \mathcal{D}_k$ and $\mathcal{D}'_1, \ldots, \mathcal{D}'_k$ be probability densities such that $\mathcal{D}_j, \mathcal{D}'_j$ are $(\varepsilon, \delta)$-indistinguishable for all $j \in [k]$. Let $\mathcal{D} = (\mathcal{D}_1, \ldots, \mathcal{D}_k)$ and $\mathcal{D}' = (\mathcal{D}'_1, \ldots, \mathcal{D}'_k)$. Then for every $\delta' > 0$, $\mathcal{D}, \mathcal{D}'$ are $(\varepsilon', k\delta + \delta')$-indistinguishable for*

$$\varepsilon' = \sqrt{2k \ln(1/\delta')}\varepsilon + k\varepsilon(e^\varepsilon - 1).$$

**Lemma C.8.** *Let $\mathcal{D}_1, \ldots, \mathcal{D}_k$ and $\mathcal{D}'_1, \ldots, \mathcal{D}'_k$ denote probability distributions on a space $\mathcal{X}$. Suppose that for all $j \in [k]$, $\mathcal{D}_j$ and $\mathcal{D}'_j$ are $(\varepsilon, \delta)$-indistinguishable. Let $w = (w_1, \ldots, w_k)$ be a probability vector, i.e. $w_j \geq 0$ for $j \in [k]$ and $\sum_{j \in [k]} w_j = 1$. Then the probability distributions $\sum_{j \in [k]} w_j \mathcal{D}_j$ and $\sum_{j \in [k]} w_j \mathcal{D}'_j$ are $(\varepsilon, \delta)$-indistinguishable.*

*Proof.* Let $\mathcal{D} = \sum_{j\in[k]} w_j \mathcal{D}_j$ and $\mathcal{D}' = \sum_{j\in[k]} w_j \mathcal{D}'_j$. Fix a set $S \subseteq \mathcal{X}$. Then

$$\mathbb{P}_{x\sim\mathcal{D}}[x \in S] = \sum_{j=1}^k w_j \mathbb{P}_{x\sim\mathcal{D}_j}[x \in S] \leq \sum_{j=1}^k w_j \left[e^\varepsilon \cdot \mathbb{P}_{x\in\mathcal{D}'_j}[x \in S] + \delta\right] = e^\varepsilon \cdot \mathbb{P}_{x\sim\mathcal{D}'}[x \in S] + \delta,$$

as required. □

**Lemma C.9** ([AL22], Lemma 2.10). *Let $\mathcal{D}_1, \mathcal{D}_2$ be continuous distributions defined on $\mathbb{R}^d$. If*

$$\mathbb{P}_{Y\sim\mathcal{D}_1}\left[\mathcal{L}_{\mathcal{D}_1\|\mathcal{D}_2}(Y) \geq \varepsilon\right] \leq \delta \quad \text{and} \quad \mathbb{P}_{Y\sim\mathcal{D}_2}\left[\mathcal{L}_{\mathcal{D}_2\|\mathcal{D}_1}(Y) \geq \varepsilon\right] \leq \delta$$

*then $\mathcal{D}_1, \mathcal{D}_2$ are $(\varepsilon, \delta)$-indistinguishable.*

# D Missing Proofs from Section 2

## D.1 Proof of Theorem 2.3

*Proof of Theorem 2.3.* **Proof of privacy.** Let $D$ and $D'$ be two neighbouring datasets and let $\mathcal{A}$ denote the non-private algorithm specified in Algorithm 1. Note that the $Q$ computed in Line 4 has sensitivity less than $\frac{2}{t}$. Since we use the Truncated Laplace mechanism in Line 7, we have (by Theorem C.6)

$$\mathbb{P}\left[\mathcal{A}(D) = \perp\right] \leq e^\varepsilon \mathbb{P}\left[\mathcal{A}(D') = \perp\right] + \delta \tag{D.1}$$

We now show that for any $T \subseteq \mathcal{Y}$, we have

$$\mathbb{P}\left[\mathcal{A}(D) \in T\right] \leq e^{2\varepsilon} \mathbb{P}\left[\mathcal{A}(D') \in T\right] + 3e^\varepsilon \delta \quad \text{and} \tag{D.2}$$

$$\mathbb{P}\left[\mathcal{A}(D) \in T \cup \{\perp\}\right] \leq e^{2\varepsilon} \mathbb{P}\left[\mathcal{A}(D') \in T \cup \{\perp\}\right] + 4e^\varepsilon \delta \tag{D.3}$$

which establishes that Algorithm 1 is $(\varepsilon, \delta)$-DP. To this end, we consider two different cases.

**Case 1:** $Q < 0.8$. In this case, $\widetilde{Q} < 0.8 + \frac{2}{t\varepsilon}\ln\left(1 + \frac{e^\varepsilon - 1}{2\delta}\right)$ with probability 1 so $\mathbb{P}\left[\mathcal{A}(D) = \perp\right] = 1$. We now verify that Eq. (D.2) and Eq. (D.3) hold. For any $T \subseteq \mathcal{Y}$, we have $\mathbb{P}\left[\mathcal{A}(D) \in T\right] = 0$ so Eq. (D.2) is trivially satisfied. To check Eq. (D.3) holds, we apply Eq. (D.1) to see that

$$\mathbb{P}\left[\mathcal{A}(D) \in T \cup \{\perp\}\right] = \mathbb{P}\left[\mathcal{A}(D) = \perp\right] \leq e^\varepsilon \mathbb{P}\left[\mathcal{A}(D') = \perp\right] + \delta \leq e^\varepsilon \mathbb{P}\left[\mathcal{A}(D') \in T \cup \{\perp\}\right] + \delta.$$

**Case 2:** $Q \geq 0.8$. Let $Y_1, \ldots, Y_t$ and $Y'_1, \ldots, Y'_t$ be the outputs in Line 2 assuming the dataset is $D, D'$, respectively. Let $j, j'$ be the output of Line 8 assuming the dataset is $D, D'$, respectively. Next, we show that $\text{dist}(Y_j, Y'_{j'}) \leq r$.

Let $S = \{\ell \in [t] : \text{dist}(Y_j, Y_\ell) \leq r/2z\}$ and $S' = \{\ell \in [t] : \text{dist}(Y'_{j'}, Y'_\ell) \leq r/2z\}$. We know that $|S| > 0.6t$ and $|S'| > 0.6t$ (by definition of $j$ in Line 8). By the inclusion-exclusion principle, we have $|S \cap S'| = |S| + |S'| - |S \cup S'| > 0.6t + 0.6t - t = 0.2t$. Thus, if $t \geq 5$, we have $|S \cap S'| > 1$ and since $|S \cap S'|$ is an integer, we must have $|S \cap S'| \geq 2$. Since $D, D'$ differ only in a single datapoint, there is some $\ell \in S \cap S'$ such that $Y_\ell = Y'_\ell$. Thus, we conclude that

$$\text{dist}(Y_j, Y'_{j'}) \leq \text{dist}(Y_j, Y_\ell) + \text{dist}(Y_\ell, Y'_{j'}) \leq z \cdot (r/2z + r/2z) = r,$$

where in the final inequality, we used that dist is a $z$-approximate $r$-restricted triangle inequality and that $\text{dist}(Y_j, Y_\ell), \text{dist}(Y_\ell, Y'_{j'}) \leq r$.

We are now ready to verify that Eq. (D.2) and Eq. (D.3) hold. Let $\mathcal{M}$ denote the mechanism described in Algorithm 1. Fix any $T \subseteq \mathcal{Y}$. Then we have

$$\begin{aligned}
\mathbb{P}\left[\mathcal{M}(D) \in T\right] &= \mathbb{P}\left[\mathcal{M}(D) \neq \perp\right] \mathbb{P}\left[\mathcal{B}(Y_j) \in T\right] \\
&\leq (e^{\varepsilon}\mathbb{P}\left[\mathcal{M}(D') \neq \perp\right] + \delta)(e^{\varepsilon}\mathbb{P}\left[\mathcal{B}(Y'_{j'}) \in T\right] + \delta) \\
&= (e^{2\varepsilon}\mathbb{P}\left[\mathcal{M}(D') \neq \perp\right] \mathbb{P}\left[\mathcal{B}(Y'_{j'}) \in T\right] + 2e^{\varepsilon}\delta + \delta^2 \\
&\leq e^{2\varepsilon}\mathbb{P}\left[\mathcal{M}(D) \in T\right] + 3e^{\varepsilon}\delta
\end{aligned}$$

where in first inequality we used the fact that $\mathcal{B}$ is a $(r, \varepsilon, \delta)$-masking mechanism, which satisfies Eq. (D.2). Next, we also have

$$\begin{aligned}
\mathbb{P}\left[\mathcal{M}(D) \in \{\perp\} \cup T\right] &= \mathbb{P}\left[\mathcal{M}(D) = \perp\right] + \mathbb{P}\left[\mathcal{M}(D) \in T\right] \\
&\leq e^{\varepsilon}\mathbb{P}\left[\mathcal{M}(D') = \perp\right] + \delta + e^{2\varepsilon}\mathbb{P}\left[\mathcal{M}(D') \in T\right] + 3e^{\varepsilon}\delta \\
&\leq e^{2\varepsilon}\mathbb{P}\left[\mathcal{M}(D') \in \{\perp\} \cup T\right] + 4e^{\varepsilon}\delta.
\end{aligned}$$

This completes the proof of privacy.

**Proof of utility.** We divide the proof into two parts.

1. First, we show that $\widetilde{Y}$ (the noisy output) concentrates around $Y^*$.
2. Second, we show that Algorithm 1 does not fail in Line 7.

For the first part, We know that $\mathcal{B}$ is $(\frac{\alpha}{2z}, \beta)$ concentrated. Furthermore, $\forall i \in [t]$, $Y_i$ satisfies $\text{dist}(Y^*, Y_i) < \frac{\alpha}{2z}$. we have

$$\begin{aligned}
\mathbb{P}[\text{dist}(\widetilde{Y}, Y^*) > \frac{\alpha}{2} + \frac{\alpha}{2}] &\leq \mathbb{P}[z \cdot \text{dist}(\widetilde{Y}, Y_j) + z \cdot \text{dist}(Y_j, Y^*) > \frac{\alpha}{2} + \frac{\alpha}{2}] \\
&\leq \mathbb{P}[\text{dist}(\widetilde{Y}, Y_j) + \text{dist}(Y_j, F^*) > \frac{\alpha}{2z} + \frac{\alpha}{2z}] \\
&\leq \mathbb{P}[\text{dist}(\widetilde{Y}, Y_j) > \frac{\alpha}{2z}] + \mathbb{P}[\text{dist}(Y_j, Y^*) > \frac{\alpha}{2z}] \\
&\leq \beta + 0 \leq \beta,
\end{aligned}$$

where the first inequality follows from the $r$-restricted $z$-approximate triangle inequality 3 (since $\alpha/2z < r/4z^2$ by assumption), and the first part of the last inequality follows the concentration of the masking mechanism. We get $\mathbb{P}[\text{dist}(\widetilde{Y}, Y_j) > \frac{\alpha}{2z}] = \beta$, because $\widetilde{Y}$ is just a masked version of $Y_j$. Also $\mathbb{P}[\text{dist}(Y_j, Y^*) > \frac{\alpha}{2z}] = 0$, because $Y_j$ is selected from $Y_i$'s, and none of them located in a distance larger than $\frac{\alpha}{2z}$ from $Y^*$ based on our assumption.

For the second part, we start by guaranteeing that $Q$ in Line 4 equals to 1. For that we need to ensure that for all $i, j \in [t]$, $\text{dist}(Y_i, Y_j) \leq \frac{r}{2z}$. To see this by triangle inequality, we have

$$\text{dist}(Y_i, Y_j) \leq z \cdot (\text{dist}(Y_i, Y^*) + \text{dist}(Y^*, Y_j))$$

since $\text{dist}(Y_i, Y^*), \text{dist}(Y^*, Y_j) \leq \frac{r}{4z^2}$. So we conclude that $Q = 1$.

Now we need to show that $\widetilde{Q} \leq 0.9$. From Line 6, $\widetilde{Q} = Q + Z$. Therefore it is enough to show $|Z| \leq 0.1$. We know that from Definition C.5 $|Z| \leq \frac{2}{t\varepsilon} \ln\left(1 + \frac{e^{\varepsilon}-1}{2\delta}\right)$. By the assumption that $t \geq \frac{20}{\varepsilon} \ln\left(1 + \frac{e^{\varepsilon}-1}{2\delta}\right)$ we conclude that $\widetilde{Q} \leq 0.9$, so the Algorithm 1 does not fail in Line 7. □

# E Missing Proofs from Section 3

## E.1 Proof of Lemma 3.2

*Proof of Lemma 3.2.* The plan is to reduce the problem of computing $\text{dist}^k$ to binary search and checking if a bipartite graph has a perfect matching.

First, we compute $\text{dist}(F_i, F_j)$ for every $i, j \in [k]$. This takes time $k^2 T_{\text{dist}}$. Note that

$$\text{dist}^k((F_1, \ldots, F_k), (F'_1, \ldots, F'_k))$$

must be one of these $k^2$ values. In addition, observe that we can determine if

$$\text{dist}^k((F_1, \ldots, F_k), (F'_1, \ldots, F'_k)) \leq x$$

for any number $x$ by consider the following bipartite graph. The disjoint node sets are $\{F_1, \ldots, F_k\}$ and $\{F'_1, \ldots, F'_k\}$ and there is an edge between $F_i, F'_j$ if and only if $\text{dist}(F_i, F'_j) \leq x$. We then determine if there is a complete bipartite matching on this graph, which takes time at most $O(k^3)$ (e.g. by using the Hungarian algorithm). Thus, we can simply combine this with a binary search on the sorted values given by $\{\text{dist}(F_i, F'_j)\}_{i,j'}$ to compute $\text{dist}^k$. □

## E.2 Proof of Lemma 3.4

*Proof of Lemma 3.4.* Computing $\mathcal{B}^k_\sigma$ only requires computing $\mathcal{B}$ a total of $k$ times and finding permutation. The former takes time $O(k \cdot T_\mathcal{B})$ and the latter takes time $O(k \log k)$ (say by sampling $k$ uniform random numbers in $[0, 1]$ and then sorting). □

## E.3 Proof of Lemma 3.5

*Proof of Lemma 3.5.* First, we prove privacy. Let $F = (F_1, \ldots, F_k) \in \mathcal{F}^k$ and $F' = (F'_1, \ldots, F'_k) \in \mathcal{F}_k$ be such that $\text{dist}^k(F, F') \leq \gamma$. In other words, there exists a permutation $\pi$ such that $\text{dist}(F_i, F'_{\pi(i)}) \leq \gamma$ for all $i \in [k]$. Since $\mathcal{B}$ is a $(\gamma, \varepsilon, \delta)$-masking mechanism, we know that $\mathcal{B}(F_i), \mathcal{B}(F'_{\pi(i)})$ are $(\varepsilon, \delta)$-indistinguishable. Thus, by advanced composition (see Theorem C.7), $(\mathcal{B}(F_1), \ldots, \mathcal{B}(F_k))$ and $(\mathcal{B}(F'_{\pi(1)}), \ldots, \mathcal{B}(F'_{\pi(k)}))$ are $(\varepsilon', k\delta + \delta')$-indistinguishable with $\varepsilon'$ as stated in the lemma. Since $\mathcal{B}^k_\sigma((F'_1, \ldots, F'_k))$ has the same distribution has $\mathcal{B}^k_\sigma((F'_{\pi(1)}, \ldots, F'_{\pi(k)}))$, we conclude, using the fact that permutation preserves privacy (see Lemma C.8), that $\mathcal{B}^k_\sigma(F)$ and $\mathcal{B}^k_\sigma(F')$ are $(\varepsilon', k\delta + \delta')$-indistinguishable.

Finally, it remains to prove accuracy (i.e. that $\mathcal{B}^k_\sigma$ is $(\alpha, k\beta)$-concentrated). Indeed, given $F = (F_1, \ldots, F_k) \in \mathcal{F}^k$, we know that $\text{dist}(\mathcal{B}(F_i), F_i) \leq \alpha$ with probability at least $1 - \beta$. Thus, by a union bound $\text{dist}(\mathcal{B}(F_i), F_i) \leq \alpha$ for all $i \in [k]$ with probability at least $1 - k\beta$. We conclude that $\text{dist}(\mathcal{B}(F), F) \leq \alpha$ with probability at least $1 - k\beta$. □

## E.4 Proof of Lemma 3.6

*Proof of Lemma 3.6.* Let $F, F', F'' \in \mathcal{F}^k$. We need to show that if $\text{dist}^k(F, F') \leq r$ and $\text{dist}^k(F', F'') \leq r$ then $\text{dist}^k(F, F'') \leq z \cdot (\text{dist}^k(F, F') + \text{dist}^k(F', F''))$. To that end, let $\pi^*_1 \in \arg\min_\pi \max_{i \in [k]}(F_i, F'_{\pi(i)})$ and let $\pi^*_2 \in \arg\min_\pi \max_{\in [k]}(F'_{\pi^*_1(i)}, F''_{\pi(i)})$. Since dist satisfies $r$-restricted $z$-approximate triangle inequality and for any $i$, $\mathrm{dist}(F_i, F'_{\pi_1^*(i)}), \mathrm{dist}(F'_{\pi_1^*(i)}, F''_{\pi_2^*(i)}) \leq r$, we have

$$\mathrm{dist}(F_i, F''_{\pi_2^*(i)}) \leq z \cdot \left( \mathrm{dist}(F_i, F'_{\pi_1^*(i)}) + \mathrm{dist}(F'_{\pi_1^*(i)}, F''_{\pi_2^*(i)}) \right)$$
$$\leq z \cdot \left( \mathrm{dist}^k(F, F') + \mathrm{dist}^k(F', F'') \right).$$

In particular

$$\mathrm{dist}^k(F, F'') \leq \max_{i \in [k]} \mathrm{dist}(F_i, F''_{\pi_2^*(i)}) \leq z \cdot \left( \mathrm{dist}^k(F, F') + \mathrm{dist}^k(F', F'') \right), \tag{E.1}$$

as required. $\square$

# F Proof of Lemma 4.2

This section is dedicated to proving Lemma 4.2. In particular, we will introduce the masking mechanism $\mathcal{B}_{\mathrm{COMP}}(w, \mu, \Sigma)$ that satisfies the conditions of Lemma 4.2. In order to add noise to a Gaussian component $(w_i, \mu_i, \Sigma_i)$ we perform a number of steps:

1. In Subsection F.1, we discuss how to noise the mixing weight of a single component. This is the most straightforward as we can simply use the Gaussian mechanism.
2. In Subsection F.2, we discuss how to noise the mean of a single component. To do this, we use an empirically re-scaled Gaussian mechanism where the empirical covariance matrix is used to shape the noise that we add to the mean. This is somewhat similar to the empirically re-scaled Gaussian mechanism used by Brown et al. [BGSUZ21].
3. In Subsection F.3, we discuss how to noise the covariance matrix of a single component. To do this, we use the noising mechanism described in Ashtiani & Liaw [AL22].

## F.1 Noising the Mixing Weights

For noising the weights, we simply use the Gaussian mechanism. Let $\mathcal{R}_{\mathrm{w}}(w, \eta) = \max(0, w + \eta g)$ where $g \sim \mathcal{N}(0, 1)$ and $w, \eta \in \mathbb{R}$.

**Lemma F.1.** Let $\alpha, \beta, \delta > 0$, $\eta = \frac{\alpha}{\sqrt{2 + 2\ln(1/\beta)}}$, and $\gamma \leq \frac{\alpha \varepsilon}{2\sqrt{2}\ln(2/\delta)\sqrt{1+\ln(1/\beta)}}$.

1. Let $w_1, w_2 \in \mathbb{R}$. If $|w_1 - w_2| \leq \gamma$ then $\mathcal{R}_{\mathrm{w}}(w_1, \eta)$ and $\mathcal{R}_{\mathrm{w}}(w_2, \eta)$ are $(\varepsilon, \delta)$-indistinguishable.
2. Let $w \in \mathbb{R}$. Then $|\mathcal{R}_{\mathrm{w}}(w, \eta) - w| \leq \alpha$ with probability at least $1 - \beta$.

*Proof.* The first item is simply the guarantee of the Gaussian Mechanism Theorem C.4 when substituting $\Delta_2 f, \sigma$ with $\gamma, \eta$ respectively (followed by post-processing to deal with the max). The second item follows from standard tail bounds on a Gaussian random variable (e.g., Lemma A.5). $\square$

## F.2 Noising the Mean

In this section, we prove that the mechanism $\mathcal{R}_{\mathrm{MEAN}}(\mu, \Sigma, \eta) = \mu + \eta g$ where $g \sim \mathcal{N}(0, \Sigma)$ effectively privatizes the mean.

**Lemma F.2.** Let $\alpha, \beta, \delta > 0$, $\eta = \sqrt{\frac{\alpha^2}{3(d + \ln(1/\beta))}}$ and let $\gamma \leq \min\{\frac{1}{2}, \frac{\varepsilon \alpha}{24 \ln(2/\delta)\sqrt{d+\ln(1/\beta)}}\}$. Let $\mu_1, \mu_2 \in \mathbb{R}^d$ and let $\Sigma_1, \Sigma_2$ be $d \times d$ positive-definite matrices. Suppose that

1. $\max\{\|\Sigma_1^{1/2}\Sigma_2^{-1}\Sigma_1^{1/2} - I_d\|_F, \|\Sigma_2^{1/2}\Sigma_1^{-1}\Sigma_2^{1/2} - I_d\|_F\} \leq \gamma$; and

2. $\max\{\|\mu_1 - \mu_2\|_{\Sigma_1}, \|\mu_1 - \mu_2\|_{\Sigma_2}\} \leq \gamma$.

Then $\mathcal{R}_{MEAN}(\mu_1, \Sigma_1, \eta)$ and $\mathcal{R}_{MEAN}(\mu_2, \Sigma_2, \eta)$ are $(\varepsilon, \delta)$-indistinguishable. In addition, if we let $\widetilde{\mu} = \mathcal{R}_{MEAN}(\mu, \Sigma, \eta)$ then $\|\widetilde{\mu} - \mu\|_\Sigma \leq \alpha$ with probability at least $1 - \beta$.

First, we prove a bound on the privacy loss.

**Lemma F.3.** *Let $\eta > 0$ and $\gamma \in (0, 1/2]$. Let $\mu_1, \mu_2 \in \mathbb{R}^d$ and let $\Sigma_1, \Sigma_2$ be $d \times d$ positive-definite matrices. Suppose that*

1. $\max\{\|\Sigma_1^{1/2}\Sigma_2^{-1}\Sigma_1^{1/2} - I_d\|_F, \|\Sigma_2^{1/2}\Sigma_1^{-1}\Sigma_2^{1/2} - I_d\|_F\} \leq \gamma$; and
2. $\max\{\|\mu_1 - \mu_2\|_{\Sigma_1}, \|\mu_1 - \mu_2\|_{\Sigma_2}\} \leq \gamma$.

*Let $Y \sim \mathcal{N}(\mu_1, \eta^2\Sigma_1)$ and define $\mathcal{L} \coloneqq \mathcal{L}_{\mathcal{N}(\mu_1, \eta^2\Sigma_1)\|\mathcal{N}(\mu_2, \eta^2\Sigma_2)}(Y)$. Then*

$$\mathcal{L} \leq \frac{\gamma^2}{2} + \frac{\gamma^2}{2\eta^2} + 2\gamma\sqrt{\ln(2/\delta)} + 2\gamma\ln(2/\delta) + 2\gamma\sqrt{2\ln(2/\delta)}/\eta \tag{F.1}$$

*with probability at least $1 - \delta$.*

*Proof.* We directly utilize Lemma A.6 and bound each term in Eq. (A.1). For the first term, we have, using Fact A.3 and that the eigenvalues of $\Sigma_2^{-1/2}\Sigma_1\Sigma_2^{-1/2}$ are at least $1/2$ by assumption (since $\gamma < 1/2$), we have[6]

$$D_{KL}\left(\mathcal{N}(\mu_1, \eta\Sigma_1) \parallel \mathcal{N}(\mu_2, \eta\Sigma_2)\right) \leq \frac{1}{2}\left[\|\Sigma_2^{-1/2}\Sigma_1\Sigma_2^{-1/2} - I_d\|_F^2 + (\mu_2 - \mu_1)^\top(\eta^2\Sigma_2)^{-1}(\mu_2 - \mu_1)\right]$$

$$\leq \frac{1}{2}\left[\gamma^2 + \frac{\gamma^2}{\eta^2}\right]$$

The second term in Eq. (A.1) is bounded by $2\gamma\sqrt{\ln(2/\delta)}$. The third term in Eq. (A.1) is bounded by $2\gamma\ln(2/\delta)$. Finally, the fourth term in Eq. (A.1) is bounded by $(1 + \gamma)\frac{\gamma}{\eta}\sqrt{2\ln(2/\delta)}$. □

The next lemma shows that $\mathcal{R}_{\text{MEAN}}(\mu, \Sigma, \eta)$ concentrates tightly around $\mu$ w.r.t. Mahalanobis distance.

**Lemma F.4.** *Let $\widetilde{\mu} = \mathcal{R}_{MEAN}(\mu, \Sigma, \eta)$. Then $\mathbb{P}\left[\|\widetilde{\mu} - \mu\|_\Sigma^2 \geq 3\eta^2(d + \ln(1/\beta))\right] \leq \beta$.*

*Proof.* Recall that $\widetilde{\mu} = \mu + \eta\Sigma^{1/2}g$ where $g \sim \mathcal{N}(0, I_d)$. Thus, $\|\widetilde{\mu} - \mu\|_\Sigma^2 = \eta^2\|g\|_2^2$. Applying Lemma A.5 gives that

$$\mathbb{P}\left[\|\widetilde{\mu} - \mu\|_\Sigma^2 \geq 3\eta^2(d + \ln(1/\beta))\right] = \mathbb{P}\left[\|g\|_2^2 \geq 3(d + \ln(1/\beta))\right]$$
$$\leq \mathbb{P}\left[\|g\|_2^2 \geq d + 2\sqrt{d\ln(1/\beta)} + 2\ln(1/\beta)\right]$$
$$\leq 1/\beta,$$

where in the first inequality, we used that $2\sqrt{d\ln(1/\beta)} \leq d + \ln(1/\beta)$. □

*Proof of Lemma F.2.* Note that

$$\gamma \leq \frac{\varepsilon\alpha}{24\ln(2/\delta)\sqrt{d + \ln(1/\beta)}} \leq \min\left\{\sqrt{\frac{\varepsilon}{2}}, \sqrt{\frac{\varepsilon\alpha^2}{6(d + \ln(1/\beta))}}, \frac{\varepsilon}{8\ln(2/\delta)}, \frac{\varepsilon\alpha}{24\sqrt{\ln(2/\delta)(d + \ln(1/\beta))}}\right\}$$

so the first claim follows by Lemma C.9 and plugging $\gamma$ and $\eta$ into Lemma F.3 to make each term in Eq. (F.1) is at most $\varepsilon/4$. Accuracy follows from Lemma F.4 using our choice of $\eta$. □

---
[6]Note that we use that $\Sigma_2^{-1/2}\Sigma_1\Sigma_2^{-1/2}$ and $\Sigma_1^{1/2}\Sigma_2^{-1}\Sigma_1^{1/2}$ have the same spectrum (see Fact A.9).

## F.3 Noising the Covariance Matrix

Define $\mathcal{R}_{\text{Cov}}(\Sigma, \eta) = \Sigma^{1/2}(I_d + \eta G)(I_d + \eta G)^\top \Sigma^{1/2}$ where $G \in \mathbb{R}^{d \times d}$ is a matrix with independent $\mathcal{N}(0,1)$ entries. We require the following lemma which is paraphrased from Lemma 5.1 and Lemma 5.2 in [AL22]. A proof can be found in Appendix F.5.

**Lemma F.5** ( [AL22], Lemma 5.1 & Lemma 5.2). *There are absolute constant $C_1, C_2 > 0$ such that the following holds. Let $\varepsilon, \delta, \beta \in (0,1]$ and set $\eta = \frac{\alpha}{C_1\sqrt{d}(\sqrt{d}+\sqrt{\ln(4/\beta)})}$.*

- *Suppose that $\gamma \leq \frac{\varepsilon\alpha}{C_2\sqrt{d^2(d+\ln(4/\beta))\cdot\ln(2/\delta)}}$. If $\Sigma_1, \Sigma_2$ are positive-definite $d \times d$ matrices such that*

$$\max\{\|\Sigma_1^{1/2}\Sigma_2^{-1}\Sigma_1^{1/2} - I_d\|, \|\Sigma_2^{1/2}\Sigma_1^{-1}\Sigma_2^{1/2} - I_d\|\} \leq \gamma$$

  *then $\mathcal{R}_{Cov}(\Sigma_1, \eta)$ and $\mathcal{R}_{Cov}(\Sigma_2, \eta)$ are $(\varepsilon, \delta)$-indistinguishable.*
- *Let $\widetilde{\Sigma} = \mathcal{R}_{Cov}(\Sigma, \eta)$. Then*

$$\max\left\{\|\Sigma^{-1/2}\widetilde{\Sigma}\Sigma^{-1/2} - I_d\|_F, \|\widetilde{\Sigma}^{-1/2}\Sigma\widetilde{\Sigma}^{-1/2} - I_d\|_F\right\} \leq \alpha$$

  *with probability at least $1 - \beta$.*

## F.4 Masking a Single Gaussian Component

Now we use the previous three subsections to devise a masking mechanism for masking a single component. Let $\eta_{\text{W}} = \frac{\alpha}{\sqrt{2+2\ln(1/\beta)}}$, $\eta_{\text{Mean}} = \frac{\alpha}{\sqrt{3(d+\ln(1/\beta))}}$ and $\eta_{\text{Cov}} = \frac{\alpha}{C_1(\sqrt{d}+\sqrt{\ln(4/\beta)})}$ for a sufficiently large constant $C_1$. Consider the mechanism

$$\mathcal{B}_{\text{Comp}}(w, \mu, \Sigma) = (\mathcal{R}_{\text{W}}(w, \eta_{\text{W}}), \mathcal{R}_{\text{Mean}}(\mu, \Sigma, \eta_{\text{Mean}}), \mathcal{R}_{\text{Cov}}(\Sigma, \eta_{\text{Cov}})) \tag{F.2}$$

*Proof of Lemma 4.2.* The fact that $\mathcal{B}_{\text{Comp}}$ is a $(\gamma, 3\varepsilon, 3\delta)$-masking follow from Lemma F.1, Lemma F.2, and Lemma F.5 along with basic composition. That $\mathcal{B}_{\text{Comp}}$ is $(\alpha, 3\beta)$-concentrated also follow from Lemma F.1, Lemma F.2, and Lemma F.5 along with a union bound. □

## F.5 Proof of Lemma F.5

To prove Lemma F.5, we require the following two lemmas from Ashtiani & Liaw [AL22]. Note that Lemma F.7 is slightly different than what is stated in Ashtiani & Liaw [AL22] but follows easily from the proof.

**Lemma F.6** ([AL22], Lemma 5.1). *Let $d \in \mathbb{N}, \eta > 0, \varepsilon \in (0,1], \delta \in (0,1], \gamma > 0$ and suppose that*

$$\gamma \leq \min\left\{\sqrt{\frac{\varepsilon}{2d(d+1/\eta^2)}}, \frac{\varepsilon}{8d\sqrt{\ln(2/\delta)}}, \frac{\varepsilon}{8\ln(2/\delta)}, \frac{\varepsilon\eta}{12\sqrt{d}\sqrt{\ln(2/\delta)}}\right\}. \tag{F.3}$$

*Let $\Sigma_1, \Sigma_2$ be two positive-definite $d \times d$ matrices. Suppose that*

$$\max\{\|\Sigma_1^{1/2}\Sigma_2^{-1}\Sigma_1^{1/2} - I_d\|, \|\Sigma_2^{1/2}\Sigma_1^{-1}\Sigma_2^{1/2} - I_d\|\} \leq \gamma.$$

*Define $\mathcal{R}_{Cov}(\Sigma, \eta) = \Sigma^{1/2}(I + \eta G)(I + \eta G)^\top \Sigma^{1/2}$ where $G \sim \mathbb{R}^{d \times d}$ is a matrix with independent $\mathcal{N}(0,1)$ entries. Then $\mathcal{R}_{Cov}(\Sigma_1, \eta)$ and $\mathcal{R}_{Cov}(\Sigma_2, \eta)$ are $(\varepsilon, \delta)$-indistinguishable.*

**Lemma F.7** ([AL22], Lemma 5.2). *There is a sufficiently large constant $C > 0$ such that the following holds. Let $\beta > 0$ and $\Sigma$ be a positive-definite $d \times d$ matrix and set $\eta = \frac{\alpha}{C\sqrt{d}(\sqrt{d}+\sqrt{\ln(4/\beta)})}$. If $\widetilde{\Sigma} = \mathcal{R}_{Cov}(\Sigma, \eta)$ then*

$$\max\left\{\|\Sigma^{-1/2}\widetilde{\Sigma}\Sigma^{-1/2} - I_d\|_F, \|\widetilde{\Sigma}^{-1/2}\Sigma\widetilde{\Sigma}^{-1/2} - I_d\|_F\right\} \leq \alpha$$

*with probability at least $1 - \beta$.*

**Remark F.8.** *Lemma F.7 is stated in Ashtiani & Liaw [AL22] with respect to spectral distance while we state it with respect to Frobenius distance. Thus, we scaled $\eta$ down by a factor of $\sqrt{d}$ compared to Ashtiani & Liaw [AL22].*

*Proof of Lemma F.5.* For the first assertion, it suffices to show that the inequality in Eq. (F.3) holds. Since Plugging $\eta$ into the fourth term of Eq. (F.3), we note that

$$\gamma \leq \frac{\varepsilon\alpha}{C_2\sqrt{d^2(d+\ln(4/\beta))\ln(2/\delta)}}. \tag{F.4}$$

Thus, it is clear that $\gamma$ is bounded above by the second and third terms of Eq. (F.3) provided $C_2$ is sufficiently large. Next, we prove that $\gamma$ is bounded above by the first term in Eq. (F.3). Indeed, we have

$$\eta^2 = \frac{\alpha^2}{C_1^2 d(\sqrt{d}+\sqrt{\ln(4/\beta)})^2} \geq \frac{\alpha^2}{2C_1^2 d(d+\ln(4/\beta))},$$

where in the last inequality we used the fact that $(a+b)^2 \leq 2a^2 + 2b^2$ for any real numbers $a, b$. Plugging this bound of $\eta^2$ into Eq. (F.3) and some calculations give that

$$\sqrt{\frac{\varepsilon}{2d(d+1/\eta^2)}} \geq \sqrt{\frac{\varepsilon\alpha^2}{C_3 d^2(d+\ln(4/\beta))}}, \tag{F.5}$$

for some constant $C_3 > 0$. Thus, if $C_2$ is large enough then the right side of Eq. (F.4) is upper bounded by the right side of Eq. (F.5). In particular, the smallest term in Eq. (F.3) is the fourth term. Finally, it is straightforward to check that $\gamma$ is at most the last term in Eq. (F.4) by plugging in the value of $\eta$. □

## G Missing Proofs from Section 5

### G.1 Proof of Lemma 5.1

*Proof.* Applying Lemma 3.5 for masking mixtures (with $\varepsilon, \delta$ in Lemma 3.5 replaced by $3\varepsilon, 3\delta$, respectively), we have, for every $\delta' > 0$, that $\mathcal{B}_{\text{GMM}}$ is a $(\gamma, \varepsilon', 3k\delta + \delta')$-masking mechanism where

$$\varepsilon' = 3\sqrt{2k\ln(1/\delta')}\varepsilon + 3k\varepsilon(e^{3\varepsilon} - 1).$$

a $(\gamma, 3\sqrt{2k\ln(1/\delta')}\varepsilon + 3k\varepsilon(e^{3\varepsilon} - 1), 3k\delta + \delta')$-masking mechanism. As this is true for any $\delta'$, we can take $\delta' = 3k\delta$ and applying the numeric inequality $e^x \leq 1 + 2x$, valid for $x < \ln(2)$ (see Fact A.7) to get that

$$\varepsilon' \leq 3\sqrt{2k\ln(1/3k\delta)}\varepsilon + 18k\varepsilon^2,$$

Finally, to prove the accuracy part ($\mathcal{B}_{\text{GMM}}$ is $(\alpha, 3k\beta)$-concentrated), we apply the accuracy part of Lemma 3.5 for masking mixtures which was proved by union bound for all $i \in [k]$. Also defining the distance to be the maximum between all three component parameters; weight $w$, mean $\mu$, and covariance matrix $\Sigma$.

$$\text{dist}_{\text{PARAM}}(\{(w_i, \mu_i, \Sigma_i)\}_{i\in[k]}, \{(w'_i, \mu'_i, \Sigma'_i)\}_{i\in[k]}) = \min_{\pi} \max_{i\in[k]} \text{dist}_{\text{COMP}}((w_{\pi(i)}, \mu_{\pi(i)}, \Sigma_{\pi(i)}), (w'_i, \mu'_i, \Sigma'_i)).$$

We can conclude that
$$\text{dist}_{\text{PARAM}}(\mathcal{B}_{\text{GMM}}(\{(w_i, \mu_i, \Sigma_i)\}_{i\in[k]}), \{(w_i, \mu_i, \Sigma_i)\}_{i\in[k]}) \leq \alpha$$
with probability at least $1 - 3k\beta$.

Now we have $\mathcal{B}_{\text{GMM}}$ is a $(\gamma, 3\sqrt{2k\ln(1/3k\delta)}\varepsilon + 18k\varepsilon^2, 6k\delta)$-masking mechanism with respect to $(\mathcal{F}_{\text{GMM}}, \text{dist}_{\text{PARAM}})$. Moreover, $\mathcal{B}_{\text{GMM}}$ is $(\alpha, 3k\beta)$-concentrated.

To simplify it, let $\varepsilon' < \ln(2)/3, \delta' < 1, \alpha' < 1, \beta' < 1$ be parameters. We set $\delta = \delta'/6k$, $\beta = \beta'/3k$, $\alpha = \alpha'$ and $\varepsilon = \min\left\{\frac{\varepsilon'}{6\sqrt{2k\ln(1/3k\delta)}}, \sqrt{\frac{\varepsilon'}{36k}}\right\} \geq \frac{\varepsilon'}{\sqrt{72k\ln(2/\delta')}}$. Then for sufficiently large constant $C$ such that if $\gamma \leq \frac{\varepsilon'\alpha'}{C_2\sqrt{k\ln(2/\delta')}\sqrt{d(d+\ln(12k/\beta'))}\cdot\ln(12k/\delta')}$, $\mathcal{B}_{\text{GMM}}$ is a $(\gamma, \varepsilon', \delta')$-masking mechanism that is $(\alpha', \beta')$-concentrated. This proves the claim. □

### G.2 Proof of Lemma 5.2

*Proof of Lemma 5.2.* Lemma 4.1 asserts that $\text{dist}_{\text{COMP}}$ satisfies 1-restricted $(3/2)$-approximate triangle inequality. Therefore, applying Lemma 3.6 (and recalling Definition 3.1) $\text{dist}_{\text{PARAM}}$ satisfies 1-restricted $(3/2)$-approximate triangle inequality. □

## H Missing Proofs from Section 6

*Proof of Lemma 6.3.* Recall from Definition 1.4 that $\text{dist}_{\text{GMM}}$ is defined as
$$\text{dist}_{\text{GMM}}(F, F') = \min_{\pi} \max_{i\in[k]} \max\left\{|w_i - w'_{\pi(i)}|, d_{\text{TV}}\left(\mathcal{N}(\mu_i, \Sigma_i), \mathcal{N}(\mu'_{\pi(i)}, \Sigma'_{\pi(i)})\right)\right\}$$
where $\pi$ is chosen from the set of all permutations over $[k]$. Also recall that
$$\text{dist}_{\text{PARAM}}(F, F') = \min_{\pi} \max_{i\in[k]} \text{dist}_{\text{COMP}}((w_{\pi(i)}, \mu_{\pi(i)}, \Sigma_{\pi(i)}), (w'_i, \mu'_i, \Sigma'_i)),$$
where $\text{dist}_{\text{COMP}}$ is as defined in Section 4.

By Theorem 1.8, it is straightforward to check that $\frac{1}{200}\text{dist}_{\text{PARAM}}(F, F') \leq \text{dist}_{\text{GMM}}(F, F') \leq \frac{1}{\sqrt{2}}\text{dist}_{\text{PARAM}}(F, F')$. □